\documentclass[12pt]{article}

\usepackage{setspace} 
\usepackage{amsmath}
\usepackage{graphicx}
\usepackage{natbib}
\usepackage{booktabs,siunitx}
\usepackage{multirow}
\usepackage{amsfonts}
\usepackage{amsthm}
\usepackage{hyperref}

\allowdisplaybreaks[4]
\graphicspath{{./art/}}

\def\T{{ \mathrm{\scriptscriptstyle T} }}
\newcommand{\bbeta}{\boldsymbol{\beta}}
\def\AIC{\textsc{aic}}
\def\BIC{\textsc{bic}}
\def\PIC{\textsc{pic}}

\newtheorem{theorem}{Theorem}
\newtheorem{lemma}{Lemma}
\newtheorem{remark}{Remark}
\newtheorem{proposition}{Proposition}

\newtheorem{assumption}{Assumption}

\title{Survival of the fittest Cox model: Pivotal variable selection for time-to-event data}

\author{Maxime van Cutsem\\ Department of Mathematics, University of Geneva\\ \texttt{Maxime.Vancutsem@unige.ch}\\
Sylvain Sardy\\ Department of Mathematics, University of Geneva\\ \texttt{sylvain.sardy@unige.ch}
}

\begin{document}

\maketitle

\begin{abstract}
We revisit Cox's proportional hazards model to improve variable selection in survival analysis. A square-root transformation of the partial likelihood renders the selection of the regularization parameter asymptotically pivotal: the baseline hazard and the censoring rate cancel exactly, and the way censoring is spread over time only enters through a bounded correction of order $1/\log n$. Using martingale residuals, we compute this penalty level by Monte Carlo without reordering the design matrix.
 The resulting criterion borrows from information criteria such as BIC and from penalized regression methods such as the lasso, taking the best of both. On simulated and real data, our method substantially improves upon state-of-the-art approaches used daily. 
\end{abstract}

  
\section{Introduction}
\subsection{Setting}
Survival analysis is concerned with the study of time-to-event data and censoring information. Such data arise in many fields, for example in medicine, where the event of interest may be the death of a patient, or in engineering, where the survival time may represent the lifetime of a machine component. At the time of analysis, the event of interest may not yet have been observed for every individual, a phenomenon known as censoring. In addition to event times and censoring indicators, survival studies often collect covariates that can influence the outcome. In a clinical oncology trial, for example, variables such as tumour stage, patient age, smoking status, or gene expression may have a substantial impact on survival. Identifying the most relevant covariates improves the interpretability of the model, reduces overfitting, and facilitates the detection of meaningful prognostic and predictive factors that can support clinical decision-making.

Formally, survival analysis assumes that two random variables underlie the data: the event time~$T$ and the censoring time $C$. Let $(T_1, C_1, \mathbf{X}_1), \dots, (T_n, C_n, \mathbf{X}_n)$ be independent and identically distributed, where $(\mathbf{X}_i\in\mathbb{R}^p)$ denotes the covariate vector for the $i$th individual for $i=1,\ldots,n$. The observed time and event indicator are defined as $Y_i=\min(T_i,C_i)$ and $\Delta_i=\mathbf{1}\{T_i\leq C_i\}$. The measured right-censored data consist of the realizations $(y_i,\delta_i,\mathbf{x}_i)_{i=1}^n$ of $(Y_i,\Delta_i,\mathbf{X}_i)_{i=1}^n$.

We assume non-informative censoring, namely that the event and censoring times are conditionally independent given the covariates: $T_i\perp C_i\mid\mathbf{X}_i$. We denote by $X\in\mathbb{R}^{n\times p}$ (with $i$th row $\mathbf x_i^\T$), $\mathbf{y}\in\mathbb{R}^n$, and $\boldsymbol{\delta}\in\{0,1\}^n$ the observed input matrix, time vector, and event indicator vector, and by $\mathbf Y=(Y_1,\ldots,Y_n)^\T$ and $\boldsymbol\Delta=(\Delta_1,\ldots,\Delta_n)^\T$ the corresponding random vectors. As usual in regression, we condition on the covariates: the design $X$ is treated as fixed, and the randomness comes from the event and censoring times. Let $\tau$ denote the end-of-study time.

The primary objective is to characterize the distribution of the event time $T$. A quantity of interest is the survival function $S(t\vert\mathbf{x}) := {\rm pr}(T>t\vert\mathbf{x})$ which gives the probability that an individual with covariates $\mathbf{x}$ survives beyond time $t$ and is commonly used for graphical representations. For model specification, however, it is often more convenient to work with the hazard function
\begin{equation*}
    h(t\vert\mathbf{x}):=\frac{f(t\vert\mathbf{x})}{S(t\vert\mathbf{x})}=\lim_{\Delta t\to 0}\frac{1}{\Delta t}{\rm pr}\left(t\leq T<t+\Delta t\,\vert\,T\geq t,\,\mathbf{x}\right)
\end{equation*}
which describes the instantaneous risk of experiencing the event at time $t$, given survival up to that time. Here $f(t\vert\mathbf{x})$ denotes the conditional density of $T$. While the survival function captures the long-term probability of survival, the hazard function captures the instantaneous failure rate. The two are tightly connected through
\begin{equation*}
    S(t\vert\mathbf{x})=\exp{\left(-\int_0^t h(u\vert\mathbf{x})\mathrm{d}u\right)}=:\exp\{-H(t\vert\mathbf{x})\}.
\end{equation*}

\subsection{The Cox model and Cox's partial likelihood}

The proportional hazards model for survival data, also known as the Cox model \citep{Cox1972}, assumes that the hazard at time $t$ depends on the covariates through
\begin{equation*}
    h(t\vert{\bf x})=h_0(t)\exp{({\bf x}^\T\bbeta)},
\end{equation*}
where $h_0$ is the unknown baseline hazard function and $\bbeta\in\mathbb R^p$ is the regression coefficients. The associated linear predictor $\mu_{\bbeta}(\mathbf{x})={\bf x}^\T{{\bbeta}}=\sum_{j=1}^p \beta_j x_j$ has no intercept to avoid a non-identifiability issue with the baseline hazard function $h_0$. Cox showed that the estimation of $\bbeta$ can be separated from that of $h_0(t)$ by introducing the partial likelihood. Assuming, for simplicity, that there are no ties among the observed times $y_1,\dots,y_n$, the log-partial likelihood of the observed data is
\begin{equation} \label{eq:partiallik}
l^{\rm partial}(\bbeta; X, \mathbf{y}, \boldsymbol{\delta}) = \sum_{i=1}^n \delta_i \left[ \mu_{\bbeta}(\mathbf{x}_i) - \log\left(\sum_{j: y_j \geq y_i} \exp\{\mu_{\bbeta}(\mathbf{x}_j)\}\right)\right],
\end{equation}
where the outer sum is taken only over individuals who experienced the event ($\delta_i=1$), and the inner sum ranges over the risk set at time $y_i$, that is, all subjects still under observation just prior to $y_i$. Unlike the full likelihood, the partial likelihood does not depend on $h_0(t)$ and can be maximized directly with respect to $\bbeta$. It remains valid when the censoring distribution depends on the covariates, provided censoring is non-informative. It is, in fact, a profile likelihood \citep{OnProfileLik2000}. 

\subsection{Model selection for survival analysis}\label{sct:intro_model_sel} 
The linearity of $\mu_{\bbeta}$ in Cox's model is convenient for at least three reasons. First, the negative log-partial likelihood function is convex and so easy to optimize. Second, the linear model can be interpreted as it captures the linear main effects in each input. Third, the set of influential inputs 
\begin{equation} \label{eq:S}
    {\cal S}:=\left\{j\in\{1,\ldots,p\} : x_j  \mbox{ has an impact on survival}\right\}
\end{equation}
corresponds to the indices of entries of ${\boldsymbol \beta}$ different from zero: ${\cal S}\equiv\left\{j\in\{1,\ldots,p\} :  \beta_j\neq 0\right\} $. 

The search of $\cal S$ is a central objective in survival analysis and constitutes the main focus of this paper. From a practical standpoint, determining which covariates ${\cal S}$ significantly influence the risk of death is crucial. In modern high-dimensional datasets such as genomic or imaging data, the number of potential covariates is often large and careful variable selection becomes crucial. We review three different approaches used in survival analysis to estimate ${\cal S}$.

The first approach fits the full Cox model by maximizing the partial likelihood~\eqref{eq:partiallik}, and tests each coefficient against $H_0:\beta_j=0$. The resulting per-covariate p-values are then adjusted with the Benjamini–Hochberg procedure~\citep{BenjaminiHochberg1995} to control the false discovery rate at a prescribed level, and the selected set collects the covariates whose adjusted
p-value falls below that level. This procedure inherits the asymptotics of the maximum partial-likelihood estimator and is therefore reliable only when $n\gg p$. When $p$ is comparable to or larger than $n$, the estimator becomes unstable or undefined, the
p-values are no longer trustworthy, and the nominal control of the false discovery rate is lost, precisely the regime arising in modern applications.

The second approach enforces sparsity on $\hat{\boldsymbol\beta}$ by maximizing~\eqref{eq:partiallik} over subsets of the covariates and selecting the size of the optimal subset with an information criterion such as \AIC~\citep{AkaikeIEEE73,AkaikeAIC73} or \BIC~\citep{Schw:esti:1978}, namely minimizing
\begin{equation*}
    \operatorname{BIC} = 2n\,l_n\left(\bbeta; X,\mathbf{y}, \boldsymbol{\delta}\right)+\lambda C(\bbeta),
\end{equation*}
where $l_n=-l^{\rm partial}/n$ and the complexity measure $C(\bbeta)=\#\left\{\beta_j\neq 0,\, j=1,\ldots,p\right\}$ counts the covariates used in the model. For censored survival data, the relevant sample size is not the number $n$ of subjects but the number $d=\sum_{i=1}^n \delta_i$ of uncensored observations (observed events). Hence, the \BIC~penalty is taken as $\lambda = \log d$ rather than $\log n$~\citep{BICsurvival2000} ($\lambda=2$ for \AIC). This penalty is derived from an asymptotic approximation of Bayes factors and is not designed to control false discoveries. Moreover, evaluating all $2^p$ candidate subsets is computationally prohibitive in high dimensions, so one usually relies on greedy forward selection instead, or evaluates the criterion along the solution path of the lasso described next.

The third approach is the lasso~\citep{Tibs:regr:1996,TisbhiraniSurvivalLASSO}, which restores tractability by convexifying the criterion: the discrete complexity measure is replaced by the convex $\ell_1$ penalty $C(\bbeta)=\|\bbeta\|_1$ turning subset selection into the convex optimization problem
\begin{equation*}
    \hat{\bbeta}=\arg\min_{\bbeta}l_n\!\left(\bbeta; X, \mathbf{y},\boldsymbol\delta\right)+\lambda\|\bbeta\|_1.
\end{equation*}
The $\ell_1$ penalty shrinks coefficients and sets many of them exactly to zero, so a single fit yields a sparse estimate $\hat{\mathcal{S}}$ and the whole solution path is computable even when $p\gg n$. The regularization level $\lambda$, which governs the size of the selected set, is typically chosen by cross-validation to optimize predictive performance. This choice targets prediction rather than support recovery, however, and provides no probabilistic control on the selected covariates: the cross-validated lasso is known to over-select. 

To address these limitations, \citet{PIC-SMS2026} introduced the Pivotal Information Criterion (\PIC), a general framework that borrows from information criteria such as \BIC~and from penalized regression methods such as the lasso, taking the best of both. Like the latter, it relies on a continuous sparsity-inducing penalty; like the former, its regularization level is fixed a priori rather than estimated from the data. Specifically, \PIC~provides a probabilistic choice of $\lambda$ under the null hypothesis $H_0:\bbeta=\mathbf{0}$ and thereby controls, at a prescribed level, the probability of selecting any covariate under the null, while retaining a high probability of recovering $\mathcal{S}$, as established for regression and classification tasks~\citep{SSM2025,PIC-SMS2026}. Our goal here is to improve model selection in survival analysis by properly deriving \PIC.

Extending \PIC~to Cox's model raises two difficulties absent from the settings of \citet{PIC-SMS2026}: the partial likelihood is not a sum of independent terms, and the nuisance parameter is not a location or scale parameter but involves the baseline hazard and the censoring distribution, two unknown functions. Our contributions are the following. First, we show that the square-root transform of the partial likelihood makes the zero-thresholding statistic asymptotically pivotal, and that it is essentially the only transform to do so. Second, writing the null score as the product of the design with the martingale residuals \citep{TherneauGrambschFleming1990}, we simulate the pivotal detection boundary without reordering the design matrix, and we prove that this Monte Carlo calibration achieves the prescribed level. Finally, simulations and several real datasets compare \PIC~with the three approaches above and with the \BIC~evaluated along the lasso path. 

\subsection{Notation and censoring patterns}\label{sct:notation}
We close this introduction by fixing the notation used throughout the paper and by describing the censoring mechanisms on which our method is assessed.

We write $p_0={\rm pr}(\Delta_i=1)$ for the probability that an observation is an event, so that $1-p_0$ is the censoring rate. Under the null hypothesis $H_0:\bbeta=\mathbf 0$, the law of the survival data is governed by two unknown functions, the baseline hazard $h_0$ and the survival function $S_C(t)={\rm pr}(C>t)$ of the censoring time; we gather them in the nuisance parameter $\vartheta=(h_0,S_C)$.

Our calibration does not depend on the censoring rate $1-p_0$, but it depends mildly on how censoring is spread over time (Theorem~\ref{thm:PIC}). A single number $\kappa_\vartheta$ summarizes this censoring pattern. Let $F_Y$ be the distribution function of the observed time $Y$, so that $1-F_Y(Y)$ is the fraction of observations still at risk at time $Y$, and let
\begin{equation}\label{eq:kappa}
    \kappa_\vartheta=1+E\bigl[\log\{1-F_Y(Y)\}\mid\Delta=1\bigr].
\end{equation}
If events and censored observations are spread alike over the observed times, $F_Y(Y)$ is uniform given $\Delta=1$ and $\kappa_\vartheta=0$. When the censored observations tend to be the latest ones, events occur while many subjects are still at risk and $\kappa_\vartheta>0$; when they tend to be the earliest ones, $\kappa_\vartheta<0$. For a censoring rate $1-p_0$, $\kappa_\vartheta$ always lies between $\log p_0$, attained when the censored observations are all the earliest, and $-(1-p_0)\log(1-p_0)/p_0\le1$, attained when they are all the latest. 

We consider five censoring mechanisms, each with one free parameter set to reach a target censoring rate. With $30\%$ censoring and under $H_0$,  the values of $\kappa_\vartheta$ 
 are as follows.
\begin{itemize}
\item[(i)] Random censoring: $C$ is exponential and independent of $T$. With exponential event times, this is the proportional hazards model of random censorship of \citet{KoziolGreen1976}, much used to study the Kaplan--Meier estimator \citep{ChenHollanderLangberg1982}, in which $S_C=S_T^{\xi}$ for some $\xi>0$ and $Y$ and $\Delta$ are independent; $\kappa_\vartheta=0$.
\item[(ii)] Administrative censoring: every subject enters at time zero and is followed until the end of the study, $C=\tau$, so that the censored observations are the latest ones; $\kappa_\vartheta\approx0.52$, the upper bound.
\item[(iii)] Staggered entry: subjects enter uniformly over $[0,\tau/2]$ and are followed until $\tau$, and $10\%$ of them drop out at a uniform time, as in clinical trials and cohort studies; $\kappa_\vartheta\approx0.38$.
\item[(iv)] Early censoring: censored subjects leave the study before most events occur (e.g., early withdrawal), so that the censored observations are the earliest ones; $\kappa_\vartheta\approx-0.35$. 
\item[(v)] Covariate-dependent censoring: $C$ is exponential with a rate that depends on a covariate, as when patients with a worse prognosis drop out more often. Cox's model allows it, but it is not a null model in the sense of assumption~\ref{as1} below.
\end{itemize}
Mechanisms (ii) and (iv) are thus the two extreme censoring patterns for a given rate, and (i) and (iii) lie between them. Appendix~\ref{app:3} details their implementation in our simulations.

\section{PIC for Cox's model} \label{sct:pic_cox}
Inspired by the universal threshold \citep{Dono94b}, theoretical results on thresholding estimators \citep{BuhlGeer11}, the quantile universal threshold \citep{Giacoetal17} and square-root lasso \citep{BCW11}, the Pivotal Information Criterion (adapted to our setting) is defined as
\begin{equation}\label{eq:PIC}
{\rm PIC}=\phi\!\left[l_n\!\left(\bbeta; X, \mathbf{y}, \boldsymbol{\delta}\right)\right] +\lambda C(\bbeta),
\end{equation}
where $l_n$ is a loss function (typically the negative log-likelihood normalized by the sample size $n$), $\phi$ is a univariate function, and $C\in\mathcal{C}_{\ell_1}$ is a model complexity measure drawn from the class of first-order $\ell_1$-equivalent penalties on which the \PIC~framework is built \citep{PIC-SMS2026}.
The class contains the convex $\ell_1$ penalty of the lasso, as well as the most commonly used nonconvex penalties, including SCAD~\citep{SCAD01} and MCP~\citep{MCP2010}. The statistical advantage of a nonconvex $C$ is that it shrinks the estimated nonzero coefficients less aggressively towards zero, thereby improving the estimation of the influential covariates ${\cal S}$.
The class  $\mathcal{C}_{\ell_1}$ only constrains the penalty $C$ near the origin. 
This is the condition imposed at the origin on the amenable regularizers of \citet{LohWainwright2015,LohWainwright2017}, whose further requirements away from the origin, satisfied by SCAD and MCP, yield error bounds for stationary points and support recovery without incoherence conditions. The calibration of $\lambda$ only involves the behaviour of $C$ at the origin, where its subdifferential is $[-1,1]^p$.

\PIC~sets $\lambda$ at the detection boundary $\lambda_\alpha$ of the null model. Following \citet{PIC-SMS2026}, $\lambda_\alpha$ is the smallest penalty for which $\bbeta=\mathbf 0$ is a local minimizer of~\eqref{eq:PIC} with probability at least $1-\alpha$ under $H_0:\bbeta=\mathbf 0$, for a prescribed small $\alpha$. It is the upper $\alpha$-quantile
\begin{equation}\label{eq:lambda_alpha}
    F_\Lambda^{-1}(1-\alpha)=:\lambda_\alpha
\end{equation}
of $\Lambda=\lambda_0(X,\mathbf Y_0,\boldsymbol\Delta_0)$, where $\lambda_0$ is the zero-thresholding function of Theorem~\ref{thm:PIC}(i) below and $(\mathbf Y_0,\boldsymbol\Delta_0)$ are observed times and event indicators generated under $H_0$. The law of $\Lambda$ depends a priori on the nuisance parameter $\vartheta=(h_0,S_C)$ of Section~\ref{sct:notation}, and therefore cannot be employed as such.
The role of the transformation $\phi$ is to make $\Lambda$ asymptotically pivotal, that is, independent of $\vartheta=(h_0,S_C)$.
In that case 
 $\lambda_\alpha$ it called the pivotal detection boundary $\lambda_\alpha^{\rm PDB}$, that we use to retrieve ${\cal S}$.

\subsection{Zero-thresholding function and pivotality}\label{subsct:zt}
Cox's model has no intercept and the partial likelihood does not involve $h_0$, so the loss in~\eqref{eq:PIC} has no nuisance parameter to fit under the null model, and the zero-thresholding function follows from Theorem~4 of \citet{PIC-SMS2026} as soon as $\phi\circ l_n$ is twice differentiable at the origin. Before specifying in the following theorem the choice of $(\phi,\,l_n)$ that yields the pivotal criterion for survival analysis, we state its conditions; assumptions~\ref{as2} and~\ref{as3} are Conditions~1 and~2 of \citet{PIC-SMS2026}.

\begin{assumption}\label{as1}
Under $H_0$, the pairs $(T_i,C_i)$ are independent and identically distributed, with a law that does not depend on $\mathbf x_i$. The times $T$ and $C$ have continuous distributions, the baseline hazard $h_0$ is bounded on
$[0,\tau]$, and ${\rm pr}(Y_i\ge\tau)\ge k_\tau>0$; observations are censored at $\tau$.
\end{assumption}

\begin{assumption}\label{as2}
The vectors of variables $\mathbf{x}_1, \ldots, \mathbf{x}_n \in \mathbb{R}^p$ in $X$ are fixed, uniformly bounded $\left|x_{i j}\right| \leq B_n$,  centered $\sum_{i=1}^n \mathbf{x}_i=\mathbf{0}$ and $\sum_{i}x_{ij}^2=n$ for $j=1,\ldots,p$.
\end{assumption}

\begin{assumption}\label{as3}
The number of variables $p$ is allowed to tend to infinity with $n$ such that $B_n^4(\log (p n))^7= O\left(n^{1-\nu}\right)$ for some $\nu>0$.
\end{assumption}

Assumption~\ref{as1} defines the null model as the pure-noise model of \citet{PIC-SMS2026}: under $H_0$, neither the event time nor the censoring time depends on the covariates, so that the law of $(\mathbf Y_0,\boldsymbol\Delta_0)$ does not depend on $X$. This is stronger than $\bbeta=\mathbf 0$, since Cox's model allows the censoring distribution to depend on the covariates. It is a choice of reference mechanism for the calibration rather than an assumption on the data; Section~\ref{subsct:robust} studies covariate-dependent censoring numerically.

Under $H_0$, $\Lambda$ depends on the survival data only through the ranks of the observed times and the event indicators. It is therefore invariant to any common increasing transformation of the time scale and, in particular, does not depend on $h_0$ alone: its law is determined by the censoring rate and the censoring pattern of Section~\ref{sct:notation}.

Theorem~\ref{thm:PIC} below states that $\Lambda$ is asymptotically pivotal in the sense that there exist distribution functions~$G_n$, depending on the input matrix $X$ only, such that $\sup_{t\in\mathbb R}|{\rm pr}_\vartheta(\Lambda\le t)-G_n(t)|\to0$ for every fixed~$\vartheta$. 

\begin{theorem}\label{thm:PIC}
Consider the pivotal information criterion \eqref{eq:PIC} with the square-root transform $\phi(u)=u^{1/2}$, Cox's minus-log-partial likelihood $l_n=-l^{\rm partial}/n$ for loss and complexity measure $C\in\mathcal{C}_{\ell_1}$. Then:
\begin{enumerate}
\item[\rm(i)] $\hat\bbeta=\mathbf{0}$ is a local minimizer of~\eqref{eq:PIC} if $\lambda>\lambda_0$ and is not if $\lambda<\lambda_0$, where the zero-thresholding function is
\begin{equation}\label{eq:lambda0}
    \lambda_0(X, \mathbf{y}, \boldsymbol{\delta})=\frac{\left\|\nabla_{\bbeta}l_n\left(\bbeta; X, \mathbf{y},\boldsymbol{\delta}\right)\big|_{\bbeta=\mathbf 0}\right\|_\infty}
        {2\{l_n\left(\boldsymbol{0}; X, \mathbf{y},\boldsymbol{\delta}\right)\}^{1/2}}.
\end{equation}
\item[\rm(ii)] Under  assumptions {\rm\ref{as1}--\ref{as3}}, let $(\mathbf Y_0,\boldsymbol\Delta_0)$ be observed times and event indicators generated under $H_0:\bbeta=\mathbf{0}$. Let $Z\sim\mathcal N(\mathbf 0,\Sigma_X)$ with $\Sigma_X=X^\T X/n$. For every nuisance parameter $\vartheta$ satisfying assumption~{\rm\ref{as1}}, with $\kappa_\vartheta$ defined in~\eqref{eq:kappa}, $\Lambda=\lambda_0(X,\mathbf Y_0,\boldsymbol\Delta_0)$ satisfies
\begin{equation}\label{eq:pivotal}
    \sup_{t\in\mathbb R}\Bigl|{\rm pr}_\vartheta\bigl[2\{n(\log n-1)\}^{1/2}\Lambda\le t\bigr]-{\rm pr}\bigl(\|Z\|_\infty\le t\bigr)\Bigr|\le\{4|\kappa_\vartheta|+o(1)\}\frac{\log(2p)}{\log n}+o(1).
\end{equation}
In particular, $\Lambda$ is asymptotically pivotal when $\log p=o(\log n)$.
\end{enumerate}
\end{theorem}

Appendix~\ref{app:1} provides a proof. Part~(ii) follows from a Gaussian approximation of the numerator of~\eqref{eq:lambda0} combined with a law-of-large-numbers argument for its denominator; the unknown probability $p_0$ of observing an event appears in both terms and cancels in the ratio. In words, the censoring rate does not matter: whether $30\%$ or $70\%$ of the observations are censored, $p_0$ cancels exactly in $\Lambda$, which is what the square-root transform achieves. Only the censoring pattern enters the bound, through $\kappa_\vartheta$, with an effect that vanishes as $1/\log n$: the normalization $\log n-1$ in~\eqref{eq:pivotal} makes it null when $\kappa_\vartheta=0$, as in the Koziol--Green model (Lemma~\ref{lem:Bn}), and the calibration of Section~\ref{subsct:mc} is conservative when $\kappa_\vartheta>0$ and liberal when $\kappa_\vartheta<0$. When the right-hand side of~\eqref{eq:pivotal} vanishes, asymptotic pivotality holds with $G_n(t)={\rm pr}[\|Z\|_\infty\le2\{n(\log n-1)\}^{1/2}t]$, and the quantile~\eqref{eq:lambda_alpha} can be computed under any convenient null mechanism, without estimating $h_0$ or the censoring distribution; otherwise, \eqref{eq:pivotal} bounds the error made by doing so. Plugging the resulting pivotal detection boundary $\lambda_\alpha^{\rm PDB}$ into~\eqref{eq:PIC} yields the PIC-Cox estimator
\begin{equation}\label{eq:pic_cox}
    \hat{\bbeta}=\arg\min_{\bbeta\in\mathbb R^p}\;
    \left\{-\frac{1}{n}\sum_{i=1}^n \delta_i \left[ \mu_{\bbeta}(\mathbf{x}_i) - \log\left(\sum_{j: y_j \geq y_i} \exp\{\mu_{\bbeta}(\mathbf{x}_j)\}\right)\right]\right\}^{1/2}
    +\lambda_\alpha^{\rm PDB}\,C(\bbeta).
\end{equation}

\begin{remark}\label{rem:phi}
The square root is  the only transform yielding pivotality. For a differentiable increasing $\phi$, the zero-thresholding function is $\lambda_0=\phi'\{l_n(\mathbf 0)\}\,\|\nabla_{\bbeta} l_n(\mathbf 0)\|_\infty$, where, by the proof of Theorem~\ref{thm:PIC}, $n^{1/2}\|\nabla_{\bbeta} l_n(\mathbf 0)\|_\infty$ is approximately distributed as $p_0^{1/2}\|Z\|_\infty$ and $l_n(\mathbf 0)\approx p_0\log n$. Pivotality with respect to $p_0\in(0,1]$ thus requires $\phi'(u)\propto u^{-1/2}$, that is, $\phi(u)=au^{1/2}+b$ with $a>0$. Since replacing $\phi$ by $a\phi+b$ multiplies $\lambda_\alpha^{\rm PDB}$ by $a$ and leaves the PIC-Cox estimator unchanged, the square root is the unique pivotal transform up to this equivalence.
\end{remark}

\subsection{Monte Carlo calibration}\label{subsct:mc}
In practice the design $X$ is fixed and $\lambda_\alpha^{\rm PDB}$ is estimated by Monte Carlo: at each of $M$ iterations we draw null data under a reference mechanism $\vartheta_0$, evaluate $\Lambda$, and report the empirical $(1-\alpha)$-quantile $\hat\lambda_\alpha^{\rm MC}$ of the $M$ draws. Each draw would naively require sorting the simulated times and reordering the rows of $X$ to rebuild the risk sets. We avoid it. Sort the observed times increasingly, let $\omega(i)$ be the index of the individual observed at the $i$th ordered time and $r_i=n-i+1$ the size of the corresponding risk set. As recalled in Appendix~\ref{app:2}, the score at the origin is
\begin{equation}\label{eq:martres}
    n\,\nabla_{\bbeta}l_n(\mathbf 0)=-X^\T\mathbf u,
    \qquad
    u_{\omega(i)}=\delta_{\omega(i)}-\sum_{j\le i}\frac{\delta_{\omega(j)}}{r_j},
\end{equation}
where $\mathbf u$ is the vector of martingale residuals of the null model: $\delta_i$ minus the Nelson--Aalen estimate of the cumulative hazard at $y_i$ \citep{TherneauGrambschFleming1990}. Under $H_0$, the ranks of the observed times define a uniform random permutation $\omega$, whose law does not depend on $X$. We take as reference $\vartheta_0$ mechanism (i) of Section~\ref{sct:notation} with $\xi=1$, in which $T$ and $C$ are independent and identically distributed: then $\Delta$ is independent of $Y$ with ${\rm pr}(\Delta=1)=1/2$, so that a null draw reduces to $\mathrm{Bernoulli}(1/2)$ coin flips $e_i$ attached to the ordered positions, $\delta_{\omega(i)}=e_i$ in~\eqref{eq:martres}, and to a uniform permutation $\omega$. The permutation acts on the length-$n$ vector $\mathbf u$, and $X$ is never reordered nor copied: the numerator of each draw is the sup-norm of the product $X^\T\mathbf u$ with the fixed matrix $X$, and the denominator $2(n\sum_i e_i\log r_i)^{1/2}$ costs $O(n)$. The total cost remains of order $Mnp$, as for the naive approach, but is spent in $M$ matrix--vector products with a fixed, contiguous matrix rather than in $M$ sorts and row permutations of $X$. In the timings of Appendix~\ref{app:2}, the residual scheme is $30$ to $250$ times faster than the naive one, the gain growing with $p$, and draws $2000$ replicates in at most $0.15$ seconds for $n\le500$ and $p\le1000$.

\begin{proposition}\label{prop:mc}
Let $\hat\lambda_\alpha^{\rm MC}$ be the empirical $(1-\alpha)$-quantile of $M$ independent coin-flip draws of $\Lambda$, independent of the data. Under assumptions~\ref{as1}--\ref{as3}, for every $\vartheta$ satisfying assumption~\ref{as1}, 
\begin{equation*}
    \bigl|{\rm pr}_\vartheta(\Lambda>\hat\lambda_\alpha^{\rm MC})-\alpha\bigr|\le\{4|\kappa_\vartheta|+o(1)\}\frac{\log(2p)}{\log n}+o(1) \quad \mbox{as } n,M\to\infty.
\end{equation*}
In particular, ${\rm pr}_\vartheta(\Lambda>\hat\lambda_\alpha^{\rm MC})\to\alpha$ when $\log p=o(\log n)$.
\end{proposition}

The proof, in Appendix~\ref{app:2}, combines Theorem~\ref{thm:PIC}(ii), applied to the data and to the coin-flip reference, with the Dvoretzky--Kiefer--Wolfowitz inequality for the $M$ draws. Since the coin-flip reference has $\kappa_{\vartheta_0}=0$, the Monte Carlo calibration adds no term to the bound of Theorem~\ref{thm:PIC}.

\begin{figure}
    \centering
    \includegraphics[width=\linewidth]{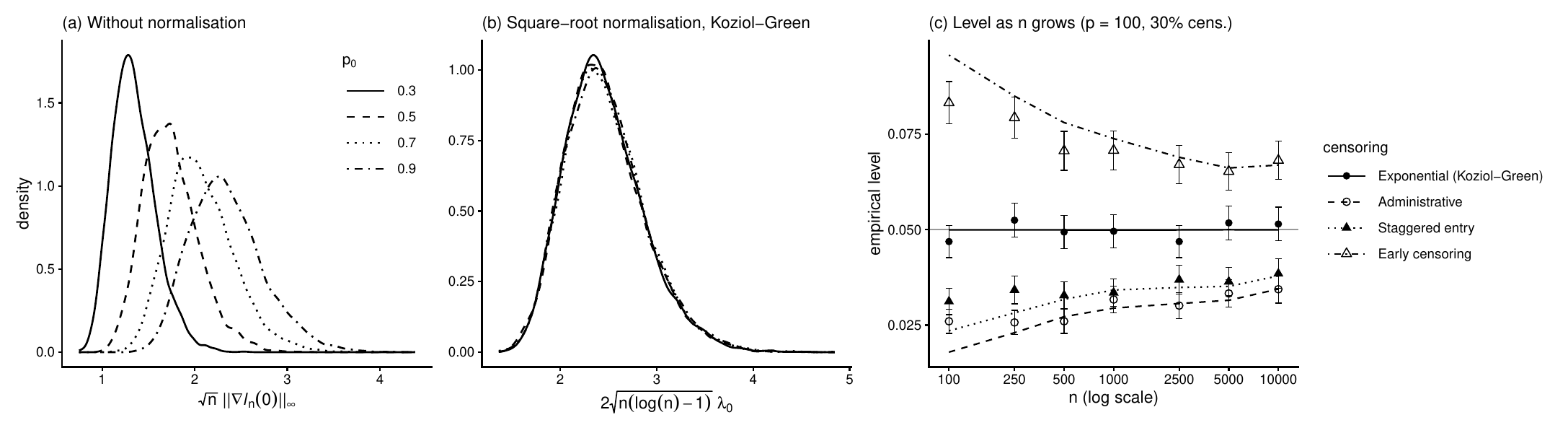}
    \caption{Empirical pivotality of the detection boundary under $H_0$. (a)--(b): fixed design
    ($n=200$, $p=50$) and $M=5000$ replicates in the Koziol--Green model at censoring levels
    $p_0\in\{0.3,0.5,0.7,0.9\}$. (a): the bare gradient, whose law depends on $p_0$ and is not pivotal.
    (b): after the square-root normalization, the densities collapse across
    censoring levels. (c): level ${\rm pr}(\hat{\mathcal S}\neq\emptyset)$ at $\alpha=0.05$ as a function of $n$ (log scale), for $p=100$ and $30\%$ censoring under four censoring mechanisms, with $10\,000$ null samples per point (bars: $\pm2$ standard errors); lines: first-order approximation obtained by rescaling the Koziol--Green statistic by $\{(\log n-1)/(\log n-1+\kappa_\vartheta)\}^{1/2}$.}
    \label{fig:pivotality}
\end{figure}

Figure~\ref{fig:pivotality} illustrates this pivotality on simulated null data. Panels (a) and (b) use a fixed
design $X\in\mathbb R^{200\times 50}$ and $M=5000$ Monte Carlo replicates drawn under
$H_0$ in the Koziol--Green model at four censoring levels, $p_0\in\{0.3,0.5,0.7,0.9\}$. Panel (a) reports the
sup-norm of the bare gradient: its law clearly depends on $p_0$, so it is not pivotal. Panel (b) reports the same quantity after the square-root normalization: the four densities collapse onto a single curve. To assess the calibration outside the Koziol--Green model, we use the five mechanisms of Section~\ref{sct:notation}, all with $30\%$ censoring. Panel (c) shows the level ${\rm pr}(\hat{\mathcal S}\neq\emptyset)$ at $\alpha=0.05$ as $n$ grows from $100$ to $10\,000$, for $p=100$ under mechanisms (i)--(iv) (mechanism (v) is used in Section~\ref{subsct:robust}), with $\lambda_\alpha^{\rm PDB}$ computed by the coin-flip scheme above. The level is exact in the Koziol--Green model at every $n$. Late censoring, administrative or with staggered entry, makes the calibration conservative, and early censoring makes it liberal; in both cases the deviation from $\alpha$ is roughly halved as $n$ grows from $100$ to $10\,000$, in line with the $1/\log n$ rate of Theorem~\ref{thm:PIC}(ii). These deviations are moderate: at $\alpha=0.05$, the level stays between $0.026$ and $0.083$, although mechanisms (ii) and (iv) nearly attain the bounds on $\kappa_\vartheta$ for $30\%$ censoring. In practice, the censoring pattern thus acts as a mild change of the nominal level. 
 The lines rescale the Koziol--Green statistic by $\{(\log n-1)/(\log n-1+\kappa_\vartheta)\}^{1/2}$, the first-order effect of the censoring pattern on the denominator of~\eqref{eq:lambda0} (Lemma~\ref{lem:Bn}), with $\kappa_\vartheta$ computed numerically for each mechanism. This first-order approximation is accurate from $n\approx500$ and overstates the deviation for smaller $n$.

\subsection{Behaviour under alternatives}\label{subsct:alt}
\textit{The null case.} Under $H_0$, every selected covariate is a false discovery, so ${\rm pr}(\hat{\mathcal S}\neq\emptyset)$ is, with the convention $0/0=0$, the false discovery rate. By Proposition~\ref{prop:mc}, it is close to $\alpha$. This is the property targeted by the calibration, and it explains why both \PIC~variants have a false discovery rate close to $\alpha$ when $s=0$ in Section~\ref{sct:simus}.

\textit{Behaviour when $s>0$.} Empirically, the calibration also behaves well under alternatives. In Section~\ref{sct:simus}, both \PIC~variants keep a false discovery rate close to zero and recover the support exactly with high probability throughout the sparse regime, far more often than the competitors. We do not attempt a theoretical analysis of the PIC-Cox estimator under alternatives, but the following argument explains why Cox's model differs from Gaussian regression in this respect.

\textit{Why Cox's model differs from Gaussian regression.} A noise covariate $j$ is selected when the stationarity condition $|\partial_j l_n(\hat\bbeta)|\le2\lambda\{l_n(\hat\bbeta)\}^{1/2}$ of~\eqref{eq:PIC} fails. The right-hand side is the penalty calibrated under $H_0$, multiplied by a normalization computed from the fitted loss. For the square-root lasso, this normalization is the residual norm, which estimates the noise level $\sigma$ whatever the true coefficients: the signal is removed from the residuals, and noise covariates face the same threshold as under $H_0$. In Cox's model, $\{l_n(\hat\bbeta)\}^{1/2}$ plays the role of the residual norm, but the loss does not separate noise from signal in the same way. With  the true coefficients $\bbeta^*$ and $\eta_i=\mathbf x_i^\T\bbeta^*$,
\begin{equation*}
    l_n(\bbeta^*)=\frac1n\sum_{i=1}^n\delta_i\log r_i
    +\frac1n\sum_{i=1}^n\delta_i\Bigl[\log\Bigl\{\frac1{r_i}\sum_{j:y_j\ge y_i}e^{\eta_j}\Bigr\}-\eta_i\Bigr],
\end{equation*}
where $r_i$ is the size of the risk set at $y_i$. The first term has the form of $l_n(\mathbf 0)$ under $H_0$ (Lemma~\ref{lem:Bn}) and plays the role of $\sigma^2$. The second term is negative for strong signals, because events occur preferentially among individuals whose risk score $\eta_i$ exceeds the risk-set average: it measures how well the signal predicts which individual fails next. Fitting the signal thus lowers the normalization below its null value, whereas the score of noise covariates 
 is barely affected. Noise covariates therefore face a threshold smaller than under $H_0$, by a factor of about $\{l_n(\hat\bbeta)/l_n(\mathbf 0)\}^{1/2}$.

\textit{Size of the effect.} The effect grows with the strength of the signal and depends on the fit. \texttt{PIC-scad}, whose fit of the active coefficients is nearly unbiased, has $l_n(\hat\bbeta)\approx l_n(\bbeta^*)$. This explains why it occasionally selects a noise covariate under strong signals, and why its probability of exact support recovery plateaus below one in Fig.~\ref{fig:phase_transition_simu} while its true positive rate equals one. The shrinkage of \texttt{PIC-lasso} keeps $l_n(\hat\bbeta)$ closer to $l_n(\mathbf 0)$, hence the threshold closer to its null value, and, with independent covariates, \texttt{PIC-lasso} remains conservative. In both cases, the signal causes a moderate rescaling of the threshold rather than a failure of the calibration, which is why the false discovery rate remains small. The two variants thus offer a trade-off. \texttt{PIC-lasso} is more conservative, but its shrinkage lowers its true positive rate as $s$ grows, so that its probability of exact support recovery collapses earlier. \texttt{PIC-scad} recovers the support for larger $s$, at the price of occasionally selecting one noise covariate too many.

\textit{Link with the lasso.} Both $l_n$ and $l_n^{1/2}$ have the same minimizers. Indeed $l_n(\bbeta)=n^{-1}\sum_i\delta_i\log[\sum_{j:y_j\ge y_i}\exp\{\mu_{\bbeta}(\mathbf x_j)-\mu_{\bbeta}(\mathbf x_i)\}]\ge0$ since each inner sum contains the term $j=i$, and $u\mapsto u^{1/2}$ is increasing. For the penalized fits, they are related as follows.

\begin{proposition}\label{prop:lasso}
Let $C=\|\cdot\|_1$. If $\hat\bbeta$ is a stationary point of~\eqref{eq:pic_cox}, then $\hat\bbeta$ minimizes the lasso criterion $l_n(\bbeta)+\tilde\lambda\|\bbeta\|_1$ for Cox's model, with $\tilde\lambda=2\lambda\{l_n(\hat\bbeta)\}^{1/2}$.
\end{proposition}

\begin{proof}
Stationarity reads $\mathbf 0\in\nabla l_n(\hat\bbeta)/\{2l_n(\hat\bbeta)^{1/2}\}+\lambda\,\partial\|\hat\bbeta\|_1$, that is, $\mathbf 0\in\nabla l_n(\hat\bbeta)+\tilde\lambda\,\partial\|\hat\bbeta\|_1$, which characterizes the minimizers of the convex lasso criterion.
\end{proof}

\texttt{PIC-lasso} is thus a point of the lasso path of \citet{TisbhiraniSurvivalLASSO}, as the square-root lasso is a point of the lasso path in linear regression. By Theorem~\ref{thm:PIC}(ii), $\lambda_\alpha^{\rm PDB}$ is of order $\{\log p/(n\log n)\}^{1/2}$, so that $\tilde\lambda$ is of order $(\log p/n)^{1/2}$ when $l_n(\hat\bbeta)$ is of order $\log n$. This is the order of penalty for which \citet{HuangCox2013} establish oracle inequalities for the lasso in Cox's model.

\section{Simulation studies and applications}\label{sct:simus}
\subsection{Setup and baselines}\label{subsct:setup}
We call \texttt{PIC-lasso} our estimator employing the lasso (i.e., convex $\ell_1$) penalty complexity measure $C$ and \texttt{PIC-scad} its nonconvex analogue. We recommend the latter, which recovers the support for larger sparsity levels; the former is more conservative under strong signals (Section~\ref{subsct:alt}). Following \citet{SCAD01}, the SCAD penalty uses $a=3.7$. The pivotal detection boundary $\lambda_\alpha^{\rm PDB}$ is computed as described in Section~\ref{subsct:mc} using significance level $\alpha=0.05$ and $2000$ Monte Carlo replications. Our method is compared against the three established baselines discussed in Section \ref{sct:intro_model_sel} and against the \BIC~evaluated along the lasso path. We recall them briefly. The first, denoted \texttt{Cox+BH}, fits the full Cox model and adjusts the obtained p-values by the Benjamini-Hochberg method~\citep{BenjaminiHochberg1995}, selecting the covariates whose adjusted p-value falls below $\alpha=0.05$. The second,
denoted \texttt{BIC}, performs greedy forward selection with the \BIC\ penalty $\lambda=\log d$ on the
number $d$ of uncensored observations rather than the total sample size $n$, which has been shown to yield more reliable results \citep{BICsurvival2000}. The third, \texttt{cv.glmnet}, is the cross-validated lasso of
\citet{TisbhiraniSurvivalLASSO} with the penalty parameter $\lambda$ selected by 10-fold cross-validation using partial-likelihood deviance, retaining the value $\lambda_{\min}$ that minimizes this criterion. The fourth, \texttt{lasso-BIC}, computes the same lasso path and retains the value of $\lambda$ minimizing $-2l^{\rm partial}\{\hat\bbeta(\lambda)\}+\hat s(\lambda)\log d$, where the partial likelihood is evaluated at the penalized coefficients and the number $\hat s(\lambda)$ of nonzero coefficients estimates the degrees of freedom of the lasso \citep{ZouHastieTibshirani2007}. Both \PIC~estimators are computed by FISTA \citep{BeckTeboulle2009}, initialized at the origin, along a decreasing path of penalties from $\lambda_0(X,\mathbf y,\boldsymbol\delta)$ down to $\lambda_\alpha^{\rm PDB}$ with warm starts; for \texttt{PIC-scad}, a final run with the SCAD penalty at $\lambda_\alpha^{\rm PDB}$ is initialized at the $\ell_1$ solution.

\subsection{Evaluation of the method by phase transition} \label{subsct:MC}
We compare the methods on simulated data through the prism of a phase transition in the probability of exact support recovery ${\rm PESR}={{\rm pr}}(\hat {\cal S}={\cal S})$ as a function of the sparsity level $s=|{\cal S}|$. This concept, originally studied by \citet{CandesTao05} and \citet{Donoho:CS:06} for compressed sensing, describes a sharp transition: ${\rm PESR}$  is high when $s$ is low, and suddenly drops to zero when $s$ gets large. The longer ${\rm PESR}$ remains near one for larger~$s$, the better the detection method. Looking at the true positive rate $\operatorname{TPR}:=E(|\hat{\cal S}\cap\cal S|/|\cal S|)$ and the false discovery rate $\operatorname{FDR}:=E(|\hat{\cal S}\cap\bar{\cal S}|/|\hat{\cal S}|)$  helps understand the reason why exact support recovery fails.

To estimate the different performance metrics, we perform $300$ independent simulation runs for each sparsity level $s\in\{0,\dots,25\}$ given a fixed standardized Gaussian design $X\in\mathbb R^{n\times p}$ with independent $\mathcal N(0,1)$ entries; 
correlated designs are considered in Section~\ref{subsct:robust}. For each run, we draw a coefficient vector $\bbeta_s$ with exactly $s$ nonzero entries, each with a $\mathcal{U}(0.75,1.5)$ magnitude, a random sign and a uniformly chosen location. Following \citet{bender2005generating}, the survival times are generated from an exponential Cox model with baseline hazard $h_0=0.1$:
\begin{equation*}
    T_i=\frac{-\log U_i}{h_0\,\exp(\mathbf x_i^\T\bbeta_s)},\qquad U_i\sim\mathcal{U}(0,1).
\end{equation*}
Censoring follows mechanism (i) of Section~\ref{sct:notation}, with about $30\%$ censoring, so that $\kappa_\vartheta=0$ under $H_0$; Appendix~\ref{app:3} repeats the experiment for $n=200$ under mechanisms (ii) and (iv), the two extreme censoring patterns.
We report two regimes that differ only by the sample size, with $p=75$: $n=100$, where $d\approx 70<p$ and the full Cox model is ill-posed, so that \texttt{Cox+BH} is omitted, and $n=200$, where $d\approx 140>p$. 

\begin{figure}
\centerline{\includegraphics[width=\linewidth]{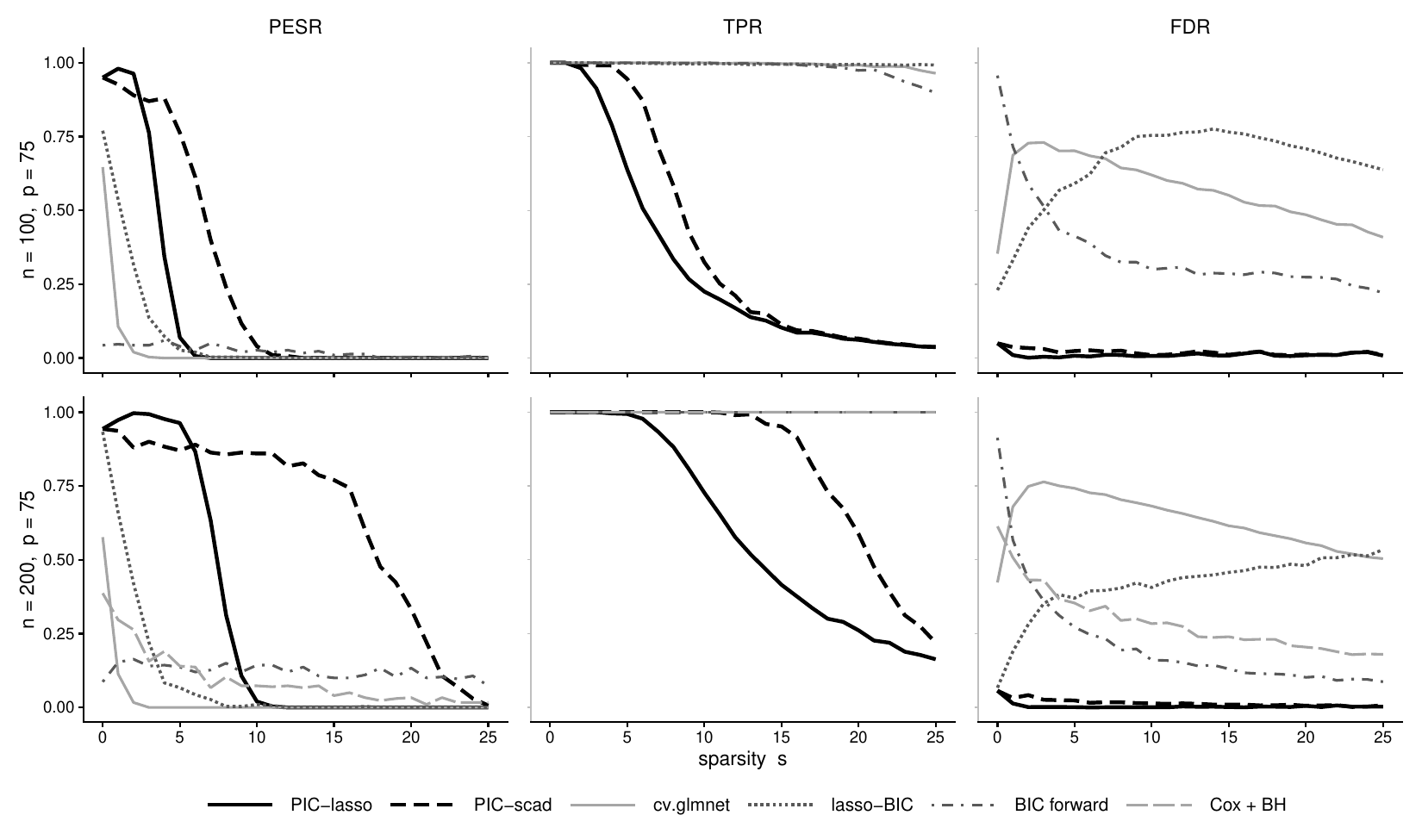}}
\caption{Phase transition of support recovery as a function of the true sparsity $s$, averaged
over $300$ Monte Carlo replicates, for $p=75$ and $|\beta_j|\sim\mathcal U(0.75,1.5)$. Columns report the probability of exact support recovery (PESR), the true positive rate (TPR), and the false discovery rate (FDR); rows correspond to $n=100$ ($d\approx70<p$) and $n=200$ ($d\approx140>p$). For $n=100$, \texttt{Cox+BH} is omitted.}
\label{fig:phase_transition_simu}
\end{figure}

Figure~\ref{fig:phase_transition_simu} presents the results.
In the null case $s=0$, every selected covariate is a false discovery. Both \PIC~variants select a covariate with probability close to $\alpha=0.05$, whereas all competitors do so more often. \texttt{Cox+BH}  and the Wald p-values of the full Cox fit are far too small.
As $s$ increases, the competitors keep a true positive rate near one, but make so many false discoveries that their probability of exact support recovery stays low. Both \PIC~variants keep a false discovery rate close to zero, and their probability of exact support recovery drops only when their true positive rate declines: this is the phase transition. Doubling $n$ roughly doubles the critical sparsity of \texttt{PIC-lasso} and more than doubles that of \texttt{PIC-scad}. The gap between the two variants is the cost of $\ell_1$ shrinkage \citep{SuBogdanCandes2017}: \texttt{PIC-lasso} shrinks every active coefficient, so that the unfitted signal accumulates with $s$ and the weakest active covariates are missed, whereas SCAD leaves large coefficients nearly unpenalized. As anticipated in Section~\ref{subsct:alt}, \texttt{PIC-scad} pays for its wider sparse regime by occasionally selecting one noise covariate, which caps its probability of exact support recovery around $0.9$, while \texttt{PIC-lasso} is the more accurate for small $s$.

Variable selection presupposes a small support, and the phase transition delineates this sparse regime; beyond it, a denser predictive method such as the lasso is more appropriate since cross-validation leads to overselection. Within it, \PIC~is the only method that combines a false discovery rate close to zero, a high true positive rate and a high probability of exact support recovery, with \texttt{PIC-scad} extending the sparse regime well beyond \texttt{PIC-lasso}.

\subsection{Censoring mechanisms and correlated designs}\label{subsct:robust}
Theorem~\ref{thm:PIC} and Fig.~\ref{fig:pivotality}(c) show that the censoring pattern affects the level of the calibration at order $1/\log n$. We now assess its effect on selection under the alternative, together with that of correlated covariates. We fix $n=200$, $p=75$ and $s=5$, a sparsity at which both \PIC~variants recover the support with high probability in Section~\ref{subsct:MC}, and draw the event times as in Section~\ref{subsct:MC}. The censoring times follow each of the five mechanisms of Section~\ref{sct:notation}, with about $30\%$ censoring (ands $60\%$ in Appendix~\ref{app:3} which details how they are simulated under the alternative). Each mechanism is crossed with an independent design and two Toeplitz designs, whose rows are independent $\mathcal N(\mathbf 0,\Sigma)$ with $\Sigma_{jk}=\rho^{|j-k|}$ and $\rho\in\{0.5,0.8\}$, standardized as in assumption~\ref{as2}. For each of the fifteen configurations we perform $100$ runs on a fixed design. Table~\ref{tab:robust} reports the average number $\hat s$ of selected covariates, the false discovery rate and the true positive rate.

\begin{table}
\centering
\footnotesize
\setlength{\tabcolsep}{3.4pt}
\begin{tabular}{lcccccc}
\toprule
Censoring & \texttt{PIC-lasso} & \texttt{PIC-scad} & \texttt{cv.glmnet} & \texttt{lasso-BIC} & \texttt{BIC} & \texttt{Cox+BH} \\
\midrule
\multicolumn{7}{l}{\textit{Independent design}} \\
\addlinespace[2pt]
Exponential & 4.9 / 0 / 99 & 5.1 / 2 / 100 & 20.6 / 74 / 100 & 8.4 / 36 / 100 & 6.9 / 25 / 100 & 8.1 / 33 / 100 \\
Administrative & 4.9 / 0 / 99 & 5.1 / 2 / 100 & 20.8 / 74 / 100 & 9.2 / 41 / 100 & 7.4 / 29 / 100 & 8.4 / 34 / 100 \\
Staggered & 5.0 / 0 / 99 & 5.1 / 2 / 100 & 22.1 / 76 / 100 & 9.1 / 40 / 100 & 7.2 / 28 / 100 & 8.5 / 35 / 100 \\
Early & 4.9 / 0 / 97 & 5.1 / 1 / 100 & 20.1 / 73 / 100 & 8.7 / 36 / 100 & 7.3 / 27 / 100 & 11.1 / 46 / 100 \\
Covariate & 5.0 / 0 / 99 & 5.1 / 2 / 100 & 21.5 / 75 / 100 & 8.7 / 38 / 100 & 7.4 / 29 / 100 & 8.9 / 36 / 100 \\
\midrule
\multicolumn{7}{l}{\textit{Toeplitz design, $\rho=0.5$}} \\
\addlinespace[2pt]
Exponential & 4.9 / 3 / 95 & 5.1 / 2 / 100 & 19.9 / 73 / 100 & 10.1 / 46 / 100 & 7.0 / 25 / 100 & 8.6 / 32 / 99 \\
Administrative & 4.9 / 3 / 94 & 5.1 / 2 / 100 & 20.9 / 74 / 100 & 10.0 / 46 / 100 & 7.0 / 25 / 100 & 8.1 / 30 / 100 \\
Staggered & 4.9 / 2 / 95 & 5.1 / 1 / 100 & 20.7 / 74 / 100 & 10.0 / 43 / 100 & 6.8 / 23 / 100 & 7.9 / 29 / 100 \\
Early & 4.9 / 3 / 94 & 5.1 / 2 / 100 & 19.1 / 72 / 100 & 10.0 / 45 / 100 & 6.8 / 22 / 100 & 11.0 / 44 / 100 \\
Covariate & 5.0 / 3 / 98 & 5.1 / 2 / 100 & 20.1 / 74 / 100 & 9.9 / 45 / 100 & 7.0 / 26 / 100 & 9.2 / 37 / 100 \\
\midrule
\multicolumn{7}{l}{\textit{Toeplitz design, $\rho=0.8$}} \\
\addlinespace[2pt]
Exponential & 5.5 / 18 / 88 & 5.0 / 6 / 94 & 19.6 / 73 / 100 & 12.4 / 57 / 98 & 6.6 / 21 / 100 & 8.4 / 35 / 92 \\
Administrative & 5.8 / 22 / 88 & 5.1 / 6 / 95 & 18.5 / 72 / 100 & 13.0 / 59 / 99 & 6.7 / 22 / 100 & 8.1 / 34 / 91 \\
Staggered & 5.4 / 20 / 85 & 5.0 / 8 / 91 & 19.2 / 73 / 100 & 12.7 / 58 / 98 & 6.7 / 21 / 100 & 8.4 / 35 / 93 \\
Early & 5.7 / 21 / 87 & 5.1 / 7 / 95 & 17.9 / 71 / 100 & 13.3 / 60 / 99 & 6.6 / 21 / 100 & 10.2 / 46 / 90 \\
Covariate & 5.4 / 16 / 86 & 5.0 / 4 / 94 & 18.7 / 72 / 100 & 11.9 / 55 / 98 & 6.6 / 20 / 100 & 9.0 / 40 / 93 \\
\bottomrule
\end{tabular}
\caption{For an independent and two Toeplitz designs, average number of selected covariates, false discovery and true positive rates in percent ($\hat s$ / FDR / TPR) at $s=5$, for $n=200$, $p=75$ and $|\beta_j|\sim\mathcal U(0.75,1.5)$, averaged over $100$ runs, under five censoring mechanisms.}
\label{tab:robust}
\end{table}

The correlation barely changes the detection boundary, $\lambda_\alpha^{\rm PDB}$ decreasing from $0.0570$ for the independent design to $0.0552$ for $\rho=0.8$. The censoring mechanism has little effect on \PIC: within each design, its rows in Table~\ref{tab:robust} are nearly identical. With independent or moderately correlated covariates, \texttt{PIC-scad} has a true positive rate close to one and a false discovery rate close to zero in all configurations, and \texttt{PIC-lasso} is as accurate, except that with $\rho=0.5$ it occasionally replaces an active covariate by a correlated noise covariate, the known behaviour of the lasso when the irrepresentable condition fails \citep{ZhaoYu2006}. Strong correlation, $\rho=0.8$, degrades both variants: \texttt{PIC-scad} misses some active covariates but keeps $\hat s$ close to $s$ and a false discovery rate below $10\%$, whereas \texttt{PIC-lasso} also selects noise covariates. The competitors are insensitive to the censoring mechanism but select many noise covariates in all configurations, and \texttt{Cox+BH} again exceeds its nominal level by a wide margin.

Appendix~\ref{app:3} reports the same experiment with $60\%$ censoring, that is, about $80$ events. The conclusions are unchanged, with three differences: \texttt{PIC-lasso} becomes more conservative, with a true positive rate of about $90\%$ without correlation; early censoring, the liberal case of Theorem~\ref{thm:PIC}, raises the false discovery rate of \texttt{PIC-scad} from at most $3\%$ to about $5\%$ for $\rho\le0.5$; and \texttt{Cox+BH}, with $p/d$ close to one, breaks down, with false discovery rates of $50\%$ or more.

\subsection{Real survival data analyses}\label{subsct:real}
To evaluate our method on real-world data, we consider several publicly available survival datasets spanning a range of sample sizes, dimensionalities, and censoring rates:
the study of survival in patients with advanced lung cancer from the North Central Cancer Treatment Group (LUNG), the German Breast Cancer Study Group (GBSG), the Mayo Clinic trial on primary biliary cirrhosis of the liver (PBC), the serum free light chain cohort of \citet{Dispenzieri2012} (FLCHAIN), the 70-gene signature dataset from the Netherlands Cancer Institute (NKI70), the Mantle Cell Lymphoma dataset (MCL), and the gene-expression profiles of patients with diffuse large B-cell lymphoma treated with CHOP or R-CHOP \citep[CHOP;][]{Lenz2008}. FLCHAIN is the largest cohort, with $70\%$ censoring, mostly administrative, the conservative case of Theorem~\ref{thm:PIC}; we remove the cause of death, known only after the event, and the decile group of the free light chains, which is derived from two other covariates. To isolate the effect of the dimension, we also augment PBC with all $136$ pairwise interactions of its $17$ centred and scaled covariates (PBC-INT): the same patients then fall in the regime $d<p$. Together, these datasets cover low- to high-dimensional settings. Prior to analysis, categorical variables are one-hot encoded and observations with missing values or a zero follow-up time are removed. Table~\ref{tab:real datasets} summarizes the main characteristics of each dataset along with their sources.

\begin{table}
	\centering
	\begin{tabular}{cccccc}
		\toprule
		\textbf{Dataset} & Size $n$ & Covariates $p$ & Number of events $d$ & Censoring & Source \\
		\midrule
		\multicolumn{6}{l}{\small Low-dimensional regime ($d> p$)} \\
		\midrule
        LUNG    & 168  & 7    & 121  & $28\%$ & \texttt{survival} \\
        GBSG    & 686  & 8    & 299  & $56\%$ & \texttt{survival} \\
        PBC     & 276  & 17   & 111  & $60\%$ & \texttt{survival} \\
        FLCHAIN & 6521 & 7    & 1959 & $70\%$ & \texttt{survival} \\
		\midrule
		\multicolumn{6}{l}{\small High-dimensional regime ($d < p$)} \\
		\midrule
        NKI70   & 144  & 75   & 48   & $67\%$ & \texttt{penalized} \\
        PBC-INT & 276  & 153  & 111  & $60\%$ & \texttt{survival} \\
        MCL     & 92   & 574  & 64   & $30\%$ & \texttt{SurvSet} \\
        CHOP    & 412  & 3833 & 163  & $60\%$ & \texttt{SurvSet} \\
		\bottomrule
	\end{tabular}
    \caption{Key characteristics of the datasets used, after removal of incomplete observations. LUNG, GBSG, PBC and FLCHAIN are from the R package \texttt{survival}, and PBC-INT adds to PBC all pairwise interactions of its covariates; the high-dimensional cohort NKI70 is from the R package \texttt{penalized}, and MCL and CHOP are from the \texttt{SurvSet} repository \citep{survset}.}
	\label{tab:real datasets}
\end{table}

Our objective is again to compare the different methods  in terms of their trade-off between model complexity (measured by the number of selected variables $\hat{s}$) and generalization performance (assessed via the C-index \citep{harrell1982} on unseen data). To this end, we draw $100$ random splits of each dataset. In each split, the data are divided into a training set ($70\%$ of the samples) and a test set (remaining $30\%$), with stratification on the censoring indicator to preserve a consistent censoring rate across both subsets. For all methods, the final predictor is obtained by refitting an unpenalized Cox model on the subset of variables selected from the training data. The \texttt{Cox+BH} baseline is omitted whenever the full Cox model is not identifiable, that is when the number of events does not exceed the number of covariates. For \texttt{BIC} forward selection, the number of forward steps is capped at $50$, and the method is omitted for CHOP, where each step would require several thousand Cox fits. Table~\ref{tab:real_results} reports, for each dataset, the median number of selected
variables $\hat{s}$ and the median test C-index over the $100$ splits.

On the low-dimensional datasets, the methods reach comparable C-indices but differ in parsimony: \PIC~selects fewer variables than \texttt{cv.glmnet} and \texttt{lasso-BIC}, at a cost of at most $0.04$ in C-index, and all methods agree on the large FLCHAIN cohort, despite its late censoring. \texttt{Cox+BH} is the exception: on PBC it retains a single variable and loses about $0.2$ in C-index, because in the full model only \texttt{bil} has a Benjamini--Hochberg adjusted p-value below $0.05$. Together with Section~\ref{subsct:MC}, this shows that \texttt{Cox+BH} is either powerless or poorly calibrated, depending on $p/d$.

In  high dimension on NKI70, MCL and CHOP, \PIC~selects one to four variables with a C-index within $0.03$ of the best, whereas \texttt{cv.glmnet} selects $6$ to $23$. PBC-INT isolates the effect of the dimension: \PIC~keeps nearly the same model as on PBC, whereas \texttt{cv.glmnet} and \texttt{BIC} forward selection inflate theirs. \texttt{lasso-BIC} is unstable when $d<p$: near a saturated model, the increase of the partial likelihood outweighs the penalty $\hat s(\lambda)\log d$, so that it selects nearly saturated models on NKI70 and MCL, but no variable on a third of the CHOP splits.

The standard deviations in Table~\ref{tab:real_results} show that the test C-index varies across splits more than most differences between methods, whereas the differences in model size are large and stable. Each baseline thus has a regime in which it fails: \texttt{cv.glmnet} systematically over-selects, \texttt{BIC} forward selection and \texttt{lasso-BIC} deteriorate once $p$ exceeds the number of events, and \texttt{Cox+BH} is restricted to $d>p$, where it is either overly conservative or poorly calibrated. \PIC~is the only method that remains both well-defined and competitive across the whole range of dimensionalities.

\begin{table}
	\centering
	\footnotesize
\begin{tabular}{llcccccc}
\toprule
\textbf{Dataset} & & \texttt{PIC-lasso} & \texttt{PIC-scad} & \texttt{cv.glmnet} & \texttt{lasso-BIC} & \texttt{BIC} & \texttt{Cox+BH} \\
\midrule
\multicolumn{8}{l}{\small Low-dimensional regime ($d> p$)} \\
\midrule
LUNG & $\hat s$ & 1 $\pm$ 0.7 & 1 $\pm$ 0.7 & 4 $\pm$ 1.8 & 2 $\pm$ 1.6 & 2 $\pm$ 0.7 & 1 $\pm$ 1.1 \\
 & C-index & 0.58 $\pm$ 0.05 & 0.57 $\pm$ 0.05 & 0.61 $\pm$ 0.05 & 0.57 $\pm$ 0.05 & 0.60 $\pm$ 0.04 & 0.55 $\pm$ 0.05 \\
\addlinespace
GBSG & $\hat s$ & 3 $\pm$ 0.7 & 3 $\pm$ 0.6 & 6 $\pm$ 0.9 & 5 $\pm$ 0.9 & 3 $\pm$ 0.7 & 3 $\pm$ 1.1 \\
 & C-index & 0.67 $\pm$ 0.02 & 0.67 $\pm$ 0.02 & 0.67 $\pm$ 0.02 & 0.67 $\pm$ 0.02 & 0.67 $\pm$ 0.02 & 0.67 $\pm$ 0.02 \\
\addlinespace
PBC & $\hat s$ & 7 $\pm$ 1.0 & 5 $\pm$ 0.9 & 9 $\pm$ 1.7 & 9 $\pm$ 2.0 & 5 $\pm$ 1.1 & 1 $\pm$ 1.9 \\
 & C-index & 0.82 $\pm$ 0.04 & 0.81 $\pm$ 0.04 & 0.83 $\pm$ 0.04 & 0.83 $\pm$ 0.04 & 0.81 $\pm$ 0.04 & 0.63 $\pm$ 0.13 \\
\addlinespace
FLCHAIN & $\hat s$ & 4 $\pm$ 0.4 & 4 $\pm$ 0.4 & 5 $\pm$ 0.5 & 5 $\pm$ 0.5 & 4 $\pm$ 0.6 & 4 $\pm$ 0.7 \\
 & C-index & 0.79 $\pm$ 0.01 & 0.79 $\pm$ 0.01 & 0.79 $\pm$ 0.01 & 0.79 $\pm$ 0.01 & 0.79 $\pm$ 0.01 & 0.79 $\pm$ 0.01 \\
\midrule
\multicolumn{8}{l}{\small High-dimensional regime ($d < p$)} \\
\midrule
NKI70 & $\hat s$ & 1 $\pm$ 0.8 & 1 $\pm$ 0.7 & 11 $\pm$ 24.3 & 57 $\pm$ 6.7 & 11 $\pm$ 5.5 & -- \\
 & C-index & 0.68 $\pm$ 0.09 & 0.66 $\pm$ 0.09 & 0.66 $\pm$ 0.08 & 0.64 $\pm$ 0.08 & 0.69 $\pm$ 0.07 & -- \\
\addlinespace
PBC-INT & $\hat s$ & 7 $\pm$ 1.4 & 6 $\pm$ 1.6 & 14 $\pm$ 5.0 & 12 $\pm$ 14.3 & 13 $\pm$ 3.3 & -- \\
 & C-index & 0.80 $\pm$ 0.04 & 0.78 $\pm$ 0.04 & 0.80 $\pm$ 0.04 & 0.80 $\pm$ 0.04 & 0.77 $\pm$ 0.04 & -- \\
\addlinespace
MCL & $\hat s$ & 1 $\pm$ 0.9 & 1 $\pm$ 0.9 & 6 $\pm$ 4.5 & 55 $\pm$ 2.8 & 23 $\pm$ 1.9 & -- \\
 & C-index & 0.68 $\pm$ 0.07 & 0.68 $\pm$ 0.07 & 0.65 $\pm$ 0.07 & 0.60 $\pm$ 0.07 & 0.59 $\pm$ 0.09 & -- \\
\addlinespace
CHOP & $\hat s$ & 4 $\pm$ 1.4 & 2 $\pm$ 1.2 & 23 $\pm$ 12.7 & 3 $\pm$ 3.0 & -- & -- \\
 & C-index & 0.63 $\pm$ 0.03 & 0.61 $\pm$ 0.04 & 0.63 $\pm$ 0.04 & 0.61 $\pm$ 0.07 & -- & -- \\
\bottomrule
\end{tabular}
	\caption{Number of selected variables $\hat{s}$ and test C-index over
	$100$ train/test splits: median $\pm$ standard deviation across the splits. \texttt{Cox+BH} is omitted in the high-dimensional regime,
	where the full Cox model is not identifiable, and \texttt{BIC} forward selection is omitted for CHOP ($p=3833$).}
	\label{tab:real_results}
\end{table}

\subsection{Primary Biliary Cirrhosis Study} \label{subsct:PBCS}
We now perform a comprehensive study of the Mayo Clinic trial on primary biliary cirrhosis (PBC) of the liver \citep{PBCdata}. The data contain $p=17$ covariates: \texttt{age} (in
years), \texttt{alb} (albumin in g/dl), \texttt{alk} (alkaline phosphatase in units/litre),
\texttt{bil} (serum bilirubin in mg/dl), \texttt{chol} (serum cholesterol in mg/dl),
\texttt{cop} (urine copper in $\mu$g/day), \texttt{plat} (platelets per cubic ml/1000),
\texttt{prot} (prothrombin time in seconds), \texttt{sgot} (liver enzyme in units/ml),
\texttt{trig} (triglycerides in mg/dl), \texttt{asc} (absence/presence of ascites),
\texttt{oed} ($0$ no oedema, $0.5$ untreated or successfully treated, $1$ unsuccessfully
treated), \texttt{hep} (absence/presence of hepatomegaly), \texttt{sex} ($0$ male, $1$
female), \texttt{spid} (absence/presence of spiders), \texttt{stage} (histological stage,
graded $1$ to $4$) and \texttt{trt} ($1$ control, $2$ treatment). The data consist of $418$ individuals, but we restrict our study to the $n=276$ observations without missing values, among which $111$ died before the end of the study. The ordinal covariates \texttt{oed} and \texttt{stage} are kept as numerical, \texttt{sex} is coded as binary, and each variable is standardized by subtracting its mean and dividing by its standard deviation.

\citet{TisbhiraniSurvivalLASSO} and \citet{AdaptiveLASSOCox} applied the lasso and the adaptive lasso to these data, retaining nine and eight variables respectively, in close agreement with the eight variables obtained by stepwise selection. A core of six variables recurs across all these analyses: \texttt{bil}, \texttt{alb}, \texttt{cop}, \texttt{age}, \texttt{oed} and \texttt{stage}. \texttt{PIC-scad} selects these six variables and \texttt{prot} (seven in total), and \texttt{PIC-lasso} further adds \texttt{asc} (eight in total); \texttt{cv.glmnet} and \texttt{lasso-BIC} retain nine variables, \texttt{BIC} eight, and \texttt{Cox+BH} only \texttt{bil}. These selections are stable around the detection boundary: varying $\lambda$ by $\pm15\%$ leaves the \texttt{PIC-lasso} model unchanged, and only \texttt{prot}, the weakest variable, leaves the \texttt{PIC-scad} model when $\lambda$ exceeds $\lambda_\alpha^{\rm PDB}$ by $2.5\%$.

Table~\ref{tab:PBC results} reports the estimated coefficients, which show the shrinkage profile of each method. 
Seven variables, all with small MLE coefficients ($|\hat\beta|\le0.12$), are excluded by every method: \texttt{trt}, \texttt{sex}, \texttt{hep}, \texttt{spid}, \texttt{chol}, \texttt{alk} and \texttt{trig}. The methods disagree only on \texttt{asc} and \texttt{sgot}, which suggests that they lie near the boundary between signal and noise.
The coefficients of \texttt{PIC-scad} have the magnitude of their full-model MLEs, whereas \texttt{PIC-lasso} shrinks several of them strongly (e.g., \texttt{age} and \texttt{stage}). \texttt{cv.glmnet} and \texttt{lasso-BIC} sit between the two and compensate by including weak predictors such as \texttt{asc} and \texttt{sgot}. \texttt{BIC} forward selection, which refits an unpenalized model on the selected subset, is closest to the MLE.

\begin{table}
	\centering
	\small
	\setlength{\tabcolsep}{4pt}
	\begin{tabular}{c|ccccccc}
\toprule
\textbf{Variable} & MLE & \texttt{PIC-lasso} & \texttt{PIC-scad} & \texttt{cv.glmnet} & \texttt{lasso-BIC} & \texttt{BIC} & \texttt{Cox+BH} \\
\midrule
trt & -0.062 & -- & -- & -- & -- & -- & -- \\
age & 0.304 & 0.068 & 0.277 & 0.166 & 0.205 & 0.330 & -- \\
sex & -0.120 & -- & -- & -- & -- & -- & -- \\
asc & 0.022 & 0.038 & -- & 0.026 & 0.021 & -- & -- \\
hep & 0.013 & -- & -- & -- & -- & -- & -- \\
spid & 0.046 & -- & -- & -- & -- & -- & -- \\
oed & 0.273 & 0.143 & 0.217 & 0.177 & 0.188 & 0.222 & -- \\
bil & 0.367 & 0.378 & 0.457 & 0.385 & 0.384 & 0.391 & 0.367 \\
chol & 0.115 & -- & -- & -- & -- & -- & -- \\
alb & -0.299 & -0.167 & -0.313 & -0.222 & -0.239 & -0.290 & -- \\
cop & 0.219 & 0.215 & 0.307 & 0.244 & 0.247 & 0.251 & -- \\
alk & 0.002 & -- & -- & -- & -- & -- & -- \\
sgot & 0.230 & -- & -- & 0.068 & 0.111 & 0.248 & -- \\
trig & -0.064 & -- & -- & -- & -- & -- & -- \\
plat & 0.084 & -- & -- & -- & -- & -- & -- \\
prot & 0.234 & 0.068 & 0.196 & 0.128 & 0.152 & 0.229 & -- \\
stage & 0.387 & 0.160 & 0.334 & 0.228 & 0.258 & 0.369 & -- \\
\bottomrule
\end{tabular}
	\caption{Estimated coefficients on the PBC dataset. MLE denotes the full (unpenalized) Cox fit; the remaining columns report the selected models for \texttt{PIC-lasso}, \texttt{PIC-scad}, \texttt{cv.glmnet}, \texttt{lasso-BIC} (penalized coefficients), \texttt{BIC} forward selection (refitted coefficients), and \texttt{Cox+BH} (full-model coefficients). A dash indicates a variable not selected.}
	\label{tab:PBC results}
\end{table}

\begin{figure}
\centerline{\includegraphics[width=\linewidth]{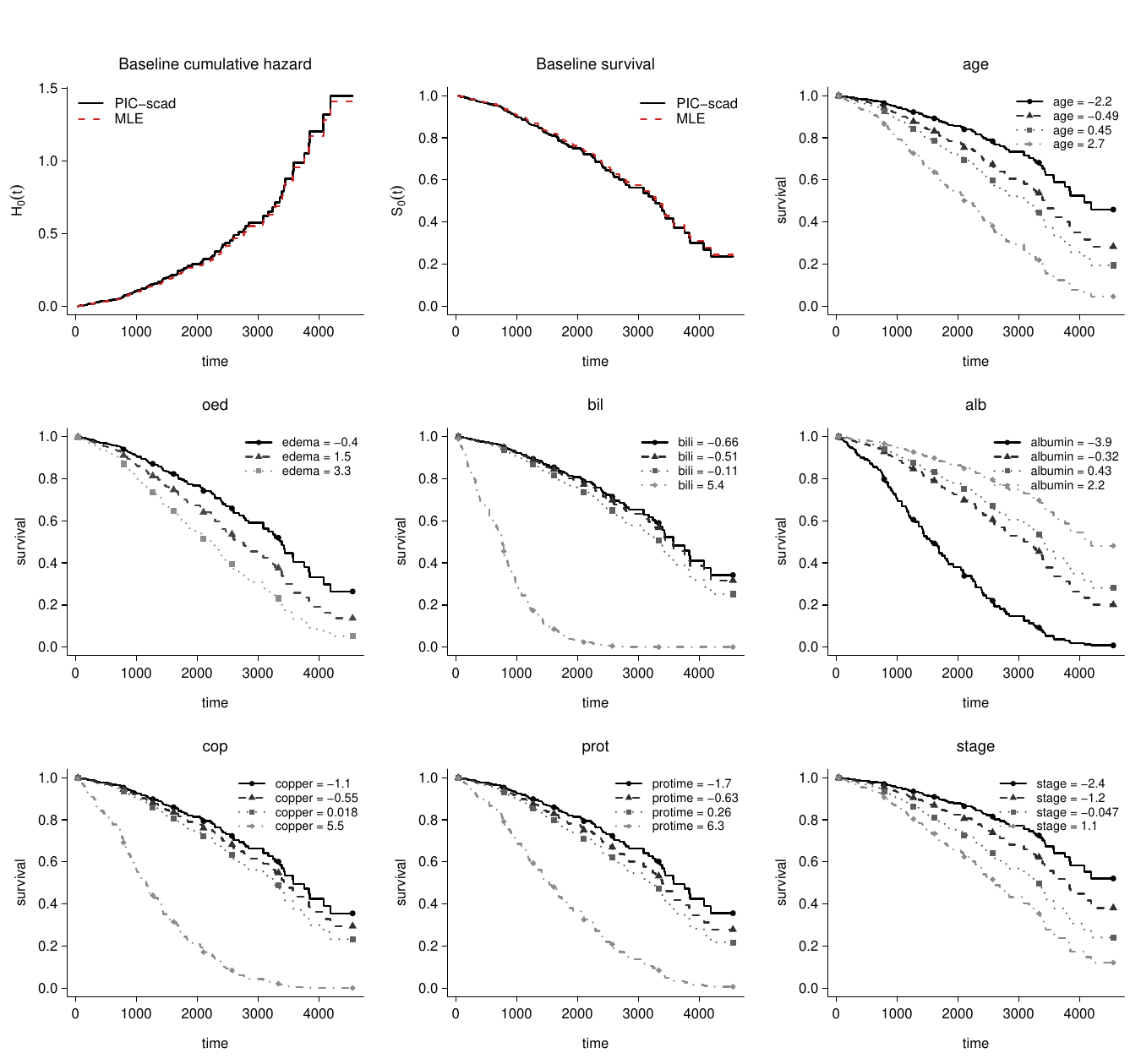}}
\caption{PIC-scad fit on the PBC dataset. The first two panels show the Breslow estimates of the baseline cumulative hazard and of the baseline survival function for the PIC-scad model (solid black) and the full MLE (dashed red). The remaining panels display the effect of each selected covariate on the survival curve, obtained by varying the covariate over representative values while holding all others at their sample mean.}
\label{fig:pbc_analysis}
\end{figure}

Figure~\ref{fig:pbc_analysis} illustrates the fitted \texttt{PIC-scad} model. Its Breslow baseline cumulative hazard and baseline survival are nearly indistinguishable from those of the full MLE, although it uses seven covariates instead of seventeen. The remaining panels display the effect of each selected covariate on the survival curve, varying it across its empirical quartiles with the other covariates fixed at their sample mean: \texttt{bil} and \texttt{stage} produce the widest spread, and \texttt{alb} is protective, as expected clinically. Such covariate-specific curves give a clinician-readable summary of the model, which the parsimony of \texttt{PIC-scad} makes possible.

\section{Conclusions} \label{sct:conclusion}
We have introduced PIC-Cox, an extension of the pivotal information criterion to the Cox
proportional-hazards model. The method applies a square-root transformation to the partial
likelihood, which removes the dependence of the zero-thresholding statistic on the censoring rate, so
that the resulting detection boundary $\lambda_\alpha^{\rm PDB}$ is asymptotically pivotal:
its law depends neither on the
baseline hazard nor on the censoring rate, but only on the input matrix and on the censoring pattern (only through a correction of relative order $1/\log n$). This pivotality is what allows the penalty level
to be calibrated once, under the null, without cross-validation and without knowledge of any
nuisance parameter. The correction is conservative when censored observations tend to be the latest ones, as in most clinical studies.

The simulation study and the real-data analyses point to a consistent picture. PIC-Cox
performs genuine variable selection with control of false discoveries under the null, and its two
variants trace a favourable complexity--accuracy trade-off: across both low- and
high-dimensional datasets, they attain a predictive C-index comparable or superior to that of
the cross-validated lasso, the BIC along the lasso path, BIC forward selection and the Benjamini--Hochberg-adjusted Cox
model, while selecting substantially fewer variables. The Benjamini--Hochberg-adjusted Cox model, the only competitor designed to control false discoveries, exceeds its nominal level by a wide margin as soon as $p/d$ is not small. The nonconvex \texttt{PIC-scad} variant
consistently improves on its convex \texttt{PIC-lasso} counterpart, mitigating the shrinkage
bias of the $\ell_1$ penalty and extending the region of exact support recovery; \texttt{PIC-lasso} is more conservative with independent covariates, since \texttt{PIC-scad} may then include one noise covariate too many, but with correlated covariates \texttt{PIC-scad} is both more accurate and less prone to false inclusions, which makes it our recommended default. In short,
PIC-Cox offers what a practitioner most often needs: a compact, well-calibrated model that
pinpoints the few covariates driving survival, with neither a tuning parameter to choose nor
a nuisance distribution to estimate.

The methodology is implemented in the R package \texttt{picreg}, which fits models following
the PIC paradigm across a range of generalized linear models, including linear regression,
binary classification, and Cox proportional-hazards regression. The package is available on
CRAN \citep{picreg} (\url{https://cran.r-project.org/package=picreg}), with documentation at
\url{https://vcmaxouuu.github.io/picreg/}.

\section*{Acknowledgement}
We thank Professor Eva Cantoni for useful comments. This work was partially supported by the Swiss National Science Foundation grant 200021E\_213166.

\appendix

\section{Proof of Theorem~\ref{thm:PIC}} \label{app:1}

\textit{Part (i).} Write $L=l_n^{1/2}$. Since $l_n(\mathbf 0)>0$, $L$ is twice continuously differentiable in a neighbourhood of the origin. Since Cox's loss has no nuisance parameter, Theorem~4 of \citet{PIC-SMS2026} applies with an empty set of nuisance parameters: the origin is a local minimizer of~\eqref{eq:PIC} if $\lambda>\|\nabla_{\bbeta} L(\mathbf 0)\|_\infty$ and is not if $\lambda<\|\nabla_{\bbeta} L(\mathbf 0)\|_\infty$. The chain rule gives the zero-thresholding function
\begin{equation*}
\lambda_0\left(X, \mathbf{y},\boldsymbol{\delta}\right)
=
\frac{
\left\|
\nabla_{\bbeta} l_n\left(\bbeta; X, \mathbf{y}, \boldsymbol{\delta}\right)
\big|_{\bbeta=\mathbf{0}}
\right\|_\infty
}{
2\{l_n\left(\mathbf{0}; X, \mathbf{y}, \boldsymbol{\delta}\right)\}^{1/2}
}.
\end{equation*}
\textit{Part (ii).}
Let $H_0: \bbeta=\mathbf{0}$ denote the null hypothesis that no covariates affect the survival time. Under $H_0$, the pairs $(Y_i,\Delta_i)_{i=1}^n$ are independent and identically distributed, with a law that does not depend on $X$. Throughout this proof, the design matrix $X$ is held fixed, and we study the random variable $\Lambda=\lambda_0(X,\mathbf Y,\boldsymbol\Delta)$.

Following \citet{coxAndersenGill}, we introduce some counting process notation. Write the counting and at-risk processes $N_i(t)=I\{Y_i\le t,\Delta_i=1\}$, $\mathcal Y_i(t)=I\{Y_i\ge t\}$, and the martingales $M_i(t)=N_i(t)-\int_0^t \mathcal Y_i(u)h_0(u)\,\mathrm du$, $M(t)=\sum_i M_i(t)$. Set the risk-set average
$\bar{\mathbf x}(u)=\sum_j \mathcal Y_j(u)\mathbf x_j/\sum_j \mathcal Y_j(u)$, the at-risk fraction
$\bar{\mathcal Y}(u)=n^{-1}\sum_j \mathcal Y_j(u)$, and $p_0={\rm pr}(\Delta_i=1)$ the probability of being uncensored under $H_0$.

The statistic can be written as
\begin{equation*}
    \Lambda = \frac{\|A_n\|_\infty}{B_n}=\frac{\|\frac{1}{n}\sum_{i=1}^n\Delta_i\left(\mathbf{x}_i-\bar{\mathbf x}(Y_i)\right)\|_\infty}{2\bigl\{\frac{1}{n}\sum_{i=1}^n \Delta_i\log{\left(\sum_{j=1}^n\mathcal Y_j(Y_i)\right)}\bigr\}^{1/2}}
\end{equation*}

\begin{lemma}\label{lem:score0sup}
Under assumptions \ref{as1}--\ref{as3}, with $Z_0\sim\mathcal N(\mathbf 0,p_0\Sigma_X)$ and $\Sigma_X=X^\T X/n$,
\begin{equation*}
    \sup_{t\in\mathbb R}\Bigl|\,
    {\rm pr}\bigl(n^{1/2}\|A_n\|_\infty\le t\bigr)
    -{\rm pr}\bigl(\|Z_0\|_\infty\le t\bigr)\Bigr|\longrightarrow 0 .
\end{equation*}
\end{lemma}
\begin{proof}
Using the counting process, $A_n=\frac{1}{n}\sum_{i=1}^n\int_0^\tau\left(\mathbf{x}_i-\bar{\mathbf{x}}(u)\right)\,\mathrm{d}N_i(u)$ and substituting the Doob–Meyer decomposition $\mathrm dN_i=\mathcal Y_ih_0\,\mathrm du+\mathrm dM_i$, the compensator term $\int_0^\tau \left(\mathbf{x}_i-\bar{\mathbf{x}}(u)\right)\mathcal Y_i(u)h_0(u)\mathrm{d}u$ vanishes upon summation over $i$ by definition of $\bar{\mathbf x}$. Hence,
\begin{align*}
    n^{1/2}A_n&=n^{-1/2}\sum_{i=1}^n \int_0^\tau\mathbf{x}_i\,\mathrm{d}M_i(u)-n^{-1/2}\int_0^\tau\bar{\mathbf{x}}(u)\,\mathrm{d}M(u)\\
    &=n^{-1/2}\sum_{i=1}^n \mathbf{x}_iM_i(\tau)-n^{-1/2}\int_0^\tau\bar{\mathbf{x}}(u)\,\mathrm{d}M(u).
\end{align*}
Let  $R_n=n^{-1/2}\int_0^\tau\bar{\mathbf x}(u)\,\mathrm dM(u)$. Under $H_0$ and \ref{as2} the
covariates are centred, $\sum_{j}\mathbf x_j=\mathbf 0$, and the law of $\mathcal Y_j(u)$ does not depend on $\mathbf x_j$, hence
the population risk-set average vanishes and
$\bar{\mathbf x}(u)$ is the centred empirical
process. By the maximal inequality for empirical processes and a union bound over the $p$ coordinates,
\begin{equation*}
    \sup_{u\in[0,\tau]}\ \|\bar{\mathbf x}(u)\|_\infty
    =O_{{\rm pr}}\!\bigl\{(\log p/n)^{1/2}\bigr\},
\end{equation*}
the analogue at $\bbeta=\mathbf 0$ of Lemma~E.1 of \citet{coxFang2017}. For $k\in\{1,\dots,p\}$, the process $t\mapsto n^{-1/2}\int_0^t\bar
x_k(u)\,\mathrm dM(u)$ is a square-integrable martingale, and by the predictable-variation
isometry, with $\mathrm d\langle M\rangle_u=\sum_{i=1}^n \mathcal Y_i(u)h_0(u)\,\mathrm du$,
\begin{equation*}
    E(R_{n,k}^2\mid X)
    =E\!\Bigl(\int_0^\tau\bar x_k(u)^2\,\bar{\mathcal Y}(u)\,h_0(u)\,\mathrm du\Bigr)
    \le\Bigl(\sup_{u\in[0,\tau]}\bar x_k(u)^2\Bigr)\mathcal H_\tau
    =O_{{\rm pr}}\!\Bigl(\tfrac{\log p}{n}\Bigr),
\end{equation*}
uniformly in $k$, where $\mathcal H_\tau=\int_0^\tau h_0<\infty$ by \ref{as1} and the last bound uses
the previous display. The jumps of $n^{-1/2}\bar x_k\,\mathrm dM$ are bounded by
$\tilde B_n/n^{1/2}$ with $\tilde B_n=B_n\{1+\mathcal H_\tau\}$, so Bernstein's inequality for
counting-process martingales \citep{vandeGeer1995}, union-bounded over the $p$ coordinates,
yields
\begin{equation*}
    \|R_n\|_\infty=O_{{\rm pr}}\!\Bigl(\tfrac{\log p}{n^{1/2}}\Bigr)
    =o_{{\rm pr}}\!\bigl((\log p)^{-1/2}\bigr),
\end{equation*}
the last step by assumption \ref{as3} since $(\log p)^{3/2}/n^{1/2}\to0$.

Let $S_n=n^{-1/2}\sum_{i=1}^n\mathbf{x}_iM_i(\tau)=n^{-1/2}\sum_{i=1}^n\mathbf v_i$, $M_i(\tau)=\Delta_i-\int_0^\tau \mathcal Y_i h_0$.  Under $H_0$, $M_i(\tau)$ depends only on $(T_i,C_i)$ which are i.i.d., hence the $\mathbf v_i=\mathbf x_i M_i(\tau)$ are independent. Moreover, $E\{M_i(\tau)\}=0$ and, under assumption~\ref{as1}, 
\begin{gather*}
E\{M_i(\tau)^2\}=E\left(\int_0^\tau \mathcal Y_i(u)h_0(u)\,du\right)=
\int_0^\tau {\rm pr}(T\ge u,\ C\ge u)h_0(u)\,du\\
=\int_0^\tau S_T(u)S_C(u)h_0(u)\,du
=\int_0^\tau f_T(u)S_C(u)\,du={\rm pr}(T\le C,\ T\le \tau)={\rm pr}(T\leq C)=p_0,
\end{gather*}
where, under $H_0$, $S_T(u)={\rm pr}(T>u)$ and $f_T=-S_T'$ are the survival and density functions of $T$, with hazard $h_0=f_T/S_T$, and $S_C$ is the survival function of $C$; the third equality uses the independence of $T$ and $C$ and the continuity of their distributions;
hence $\frac{1}{n}\sum_{i=1}^n \operatorname{cov}(\mathbf{v}_i)=p_0\Sigma_X$. Under assumption~\ref{as1},
\begin{equation*}
    |M_i(\tau)|\le|\Delta_i|+\int_0^\tau \mathcal Y_i(u)h_0(u)\,du\le 1+\mathcal H_\tau<\infty, 
\end{equation*}
and in particular $E\{M_i(\tau)^4\}<\infty$. Under assumption \ref{as3}, apply the
high-dimensional Gaussian approximation of \citet{Chernozhukov_2013} to the independent,
centred, bounded vectors $\tilde{\mathbf v}_i=(\mathbf v_i,-\mathbf v_i)\in\mathbb R^{2p}$, whose coordinates have variance $p_0$ by assumption~\ref{as2}, with $Z_0\sim\mathcal N(\mathbf 0,p_0\Sigma_X)$,
\begin{equation*}
    \rho_n:=\sup_{t\in\mathbb R}\bigl|{\rm pr}(\|S_n\|_\infty\le t)-{\rm pr}(\|Z_0\|_\infty\le t)\bigr|\longrightarrow0.
\end{equation*}

It remains to combine the two terms. The Gaussian maximum $\|Z_0\|_\infty$ is anti-concentrated: by \citet{CCK2015} there is a constant $K<\infty$ such that
\begin{equation*}
    \sup_{t\in\mathbb R}{\rm pr}\bigl(\,|\,\|Z_0\|_\infty-t\,|\le\varepsilon\,\bigr)\le K\,\varepsilon(\log p)^{1/2}
    \qquad(\varepsilon>0).
\end{equation*}
By our result on $\|R_n\|_\infty$, pick $\zeta_n\to0$ with ${\rm pr}(\|R_n\|_\infty>\zeta_n)\to0$ and
$\zeta_n(\log p)^{1/2}\to0$. On the event $\{\|R_n\|_\infty\le\zeta_n\}$ the inclusion
$\{n^{1/2}\|A_n\|_\infty\le t\}=\{\|-S_n+R_n\|_\infty\le t\}\subseteq\{\|S_n\|_\infty\le t+\zeta_n\}$
holds, hence
\begin{align*}
    {\rm pr}\bigl(n^{1/2}\|A_n\|_\infty\le t\bigr)
    &\le{\rm pr}(\|S_n\|_\infty\le t+\zeta_n)+{\rm pr}(\|R_n\|_\infty>\zeta_n)\\
    &\le{\rm pr}(\|Z_0\|_\infty\le t+\zeta_n)+\rho_n+o(1)\\
    &\le{\rm pr}(\|Z_0\|_\infty\le t)+\rho_n+K\zeta_n(\log p)^{1/2}+o(1),
\end{align*}
using the Gaussian approximation of $\|S_n\|_\infty$ then the anti-concentration bound. The
reverse inclusion
$\{\|S_n\|_\infty\le t-\zeta_n\}\subseteq\{n^{1/2}\|A_n\|_\infty\le t\}\cup\{\|R_n\|_\infty>\zeta_n\}$
gives the matching lower bound. Taking the supremum over $t\in\mathbb R$,
\begin{equation*}
    \sup_{t\in\mathbb R}\bigl|{\rm pr}(n^{1/2}\|A_n\|_\infty\le t)-{\rm pr}(\|Z_0\|_\infty\le t)\bigr|
    \le\rho_n+K\zeta_n(\log p)^{1/2}+o(1)\longrightarrow0,
\end{equation*}
which is the claim.
\end{proof}

\begin{lemma}\label{lem:Bn}
Under assumption \ref{as1}, $B_n^2=4\{p_0\log n+c_\vartheta+o_{\rm pr}(1)\}$, with $c_\vartheta=E[\Delta\log\{1-F_Y(Y)\}]=p_0(\kappa_\vartheta-1)$. Equivalently, $c_\vartheta=\int_0^1\pi_\vartheta(u)\log(1-u)\,{\rm d}u$, with $\pi_\vartheta(u)={\rm pr}_\vartheta\{\Delta=1\mid F_Y(Y)=u\}$ the probability that the observation at quantile $u$ of the observed times is an event.
\end{lemma}
\begin{proof}
One has
\begin{equation*}
    \frac{1}{n}\sum_{i=1}^n \Delta_i\log{\left(\sum_{j=1}^n I\{Y_j \ge Y_i\}\right)}= \log n\cdot \frac{\sum_{i=1}^n \Delta_i}{n} + \frac{1}{n}\sum_{i=1}^n \Delta_i\log{\left(\frac{1}{n}\sum_{j=1}^n I\{Y_j \ge Y_i\}\right)}.
\end{equation*}
Since $\Delta_i$ is binary, $n^{-1}\sum_{i=1}^n \Delta_i=p_0+O_{{\rm pr}}(n^{-1/2})$. So the first term is of order $p_0\log n + o_{\rm pr}(1)$ as $n\to\infty$. Indexing the observations by increasing observed time, with no ties, $\sum_{j=1}^n I\{Y_j \ge Y_i\}=n-i+1$ and
\begin{equation*}
    \frac{1}{n}\sum_{i=1}^n \Delta_i\log{\left(n^{-1}\sum_{j=1}^n I\{Y_j \ge Y_i\}\right)} = \frac{1}{n}\sum_{i=1}^n \Delta_i\log{\left(\frac{n-i+1}{n}\right)}=:s_n.
\end{equation*}
Since the rank of $Y_i$ is $n\hat F_Y(Y_i)$, with $\hat F_Y$ the empirical distribution function of the observed times, $s_n=n^{-1}\sum_i\Delta_i\log\{1-\hat F_Y(Y_i)+n^{-1}\}$. Fix $\varepsilon\in(0,1)$. On $\{u\le1-\varepsilon\}$, $\log(1-u)$ is bounded and uniformly continuous, so that, by the Glivenko--Cantelli theorem and the law of large numbers, the observations with $\hat F_Y(Y_i)\le1-\varepsilon$ contribute $\int_0^{1-\varepsilon}\pi_\vartheta(u)\log(1-u)\,{\rm d}u+o_{\rm pr}(1)$ to $s_n$. The others contribute at most
\begin{equation*}
    -\frac1n\sum_{i>n(1-\varepsilon)}\log\Bigl(\frac{n-i+1}{n}\Bigr)=\varepsilon\{1+\log(1/\varepsilon)\}+o(1)
\end{equation*}
in absolute value, by Stirling's approximation, and so does the remainder $\int_{1-\varepsilon}^1\pi_\vartheta(u)\log(1-u)\,{\rm d}u$ of $c_\vartheta$. Letting $\varepsilon\to0$ gives $s_n\to c_\vartheta$ in probability, which gives the result.
\end{proof}

It remains to combine Lemmas~\ref{lem:score0sup} and~\ref{lem:Bn}. We have
\begin{equation*}
    2\{n(\log n-1)\}^{1/2}\Lambda=r_nW,
    \qquad
    W=\frac{n^{1/2}\|A_n\|_\infty}{p_0^{1/2}},
    \qquad
    r_n=\frac{2\{p_0(\log n-1)\}^{1/2}}{B_n}.
\end{equation*}
By Lemma~\ref{lem:Bn}, $B_n^2/4=p_0(\log n-1)+p_0\kappa_\vartheta+o_{\rm pr}(1)$, so that $r_n=1-\kappa_\vartheta/\{2(\log n-1)\}+o_{\rm pr}(1/\log n)$. By Lemma~\ref{lem:score0sup}, the Kolmogorov distance $\rho_n'$ between $W$ and $\|Z\|_\infty$, with $Z\sim\mathcal N(\mathbf 0,\Sigma_X)$, tends to zero. Since the coordinates of $Z$ have unit variance by assumption~\ref{as2}, $E\|Z\|_\infty\le(2\log 2p)^{1/2}$, $\|Z\|_\infty-E\|Z\|_\infty$ is sub-Gaussian with variance factor one, and the anti-concentration inequality of \citet{CCK2015} gives $\sup_t{\rm pr}(|\|Z\|_\infty-t|\le\epsilon)\le4\epsilon\{(2\log 2p)^{1/2}+1\}$. Let
\begin{equation*}
    \epsilon_n=\Bigl\{\frac{|\kappa_\vartheta|}{2(\log n-1)}+\frac{\eta_n}{\log n}\Bigr\}\bigl\{(2\log 2p)^{1/2}+\ell_n\bigr\},
\end{equation*}
with $\eta_n\to0$ and $\ell_n\to\infty$ slowly, so that $\gamma_n={\rm pr}(|r_n-1|\,W>\epsilon_n)\to0$. As in the proof of Lemma~\ref{lem:score0sup},
\begin{align*}
    \sup_{t\in\mathbb R}\bigl|{\rm pr}[2\{n(\log n-1)\}^{1/2}\Lambda\le t]-{\rm pr}(\|Z\|_\infty\le t)\bigr|
    &\le\rho_n'+\gamma_n+4\epsilon_n\{(2\log 2p)^{1/2}+1\}\\
    &=\{4|\kappa_\vartheta|+o(1)\}\frac{\log(2p)}{\log n}+o(1),
\end{align*}
which is~\eqref{eq:pivotal}. The approximating distribution depends on $X$ only, and the nuisance parameter enters the bound only through $\kappa_\vartheta$, which concludes the proof of Theorem~\ref{thm:PIC}.

\section{Fast Monte Carlo evaluation of $\lambda_\alpha^{\rm PDB}$}\label{app:2}

By~\eqref{eq:lambda_alpha}, $\lambda_\alpha^{\rm PDB}$ is the upper $\alpha$-quantile of the random variable $\Lambda=\lambda_0(X,\mathbf Y_0,\boldsymbol\Delta_0)$ obtained by evaluating the zero-thresholding statistic~\eqref{eq:lambda0} on data drawn under $H_0:\bbeta=\mathbf{0}$. This appendix shows how to draw each replicate of $\Lambda$ without ever reordering or copying the design matrix $X$, and proves Proposition~\ref{prop:mc}.

\subsection*{Score at the origin and martingale residuals}\label{app:2:numerator}
Sort the observed times increasingly, $y_{\omega(1)}<\cdots<y_{\omega(n)}$, and write $\tilde{\mathbf x}_i=\mathbf x_{\omega(i)}$, $\tilde\delta_i=\delta_{\omega(i)}$ and $r_i=n-i+1$ for the size of the risk set $\{\omega(i),\ldots,\omega(n)\}$ at the $i$th ordered time. The unnormalized numerator of~\eqref{eq:lambda0} is the sup-norm of
\begin{equation*}
    \mathbf v=\sum_{i=1}^n \tilde\delta_i\Bigl(\tilde{\mathbf x}_i-\frac1{r_i}\sum_{j\ge i}\tilde{\mathbf x}_j\Bigr)
            =\sum_{i=1}^n \tilde\delta_i\tilde{\mathbf x}_i
             -\sum_{j=1}^n\tilde{\mathbf x}_j\sum_{i\le j}\frac{\tilde\delta_i}{r_i},
\end{equation*}
after exchanging the order of summation in the double sum. Relabelling gives $\mathbf v=\sum_{i=1}^n u_i\,\tilde{\mathbf x}_i=X^\T\mathbf u^{\omega}$, with
\begin{equation*}
    u_i=\tilde\delta_i-\sum_{j\le i}\frac{\tilde\delta_j}{r_j},\qquad u^{\omega}_{\omega(i)}=u_i,
\end{equation*}
that is, $\mathbf u^{\omega}$ is obtained by scattering the entries of $\mathbf u$ through $\omega$. The sum $\sum_{j\le i}\tilde\delta_j/r_j$ is the Nelson--Aalen estimate of the cumulative hazard at $y_{\omega(i)}$, so $\mathbf u^\omega$ is the vector of martingale residuals of the null model \citep{TherneauGrambschFleming1990}, and $n\nabla_{\bbeta} l_n(\mathbf 0)=-X^\T\mathbf u^\omega$. Hence $\Lambda=\|X^\T\mathbf u^\omega\|_\infty/\{2(n\sum_i\tilde\delta_i\log r_i)^{1/2}\}$, whose denominator does not involve $X$.

\subsection*{Null draws reduce to a permutation and coin flips}\label{app:2:reduction}
Under $H_0$ the pairs $(Y_i,\Delta_i)$ are independent and identically distributed, with a law that does not depend on $X$, and the observed times are almost surely distinct. The statistic $\Lambda$ depends on the survival data only through the permutation $\omega$ of the ranks and the ordered indicators $\tilde{\boldsymbol\Delta}=(\Delta_{\omega(1)},\ldots,\Delta_{\omega(n)})$, and $\omega$ is a uniform random permutation independent of $\tilde{\boldsymbol\Delta}$. Under the reference mechanism $\vartheta_0$ in which $T$ and $C$ are independent and identically distributed, the indicator $\Delta=I\{T\le C\}$ is independent of $Y=\min(T,C)$ by symmetry, with ${\rm pr}(\Delta=1)=1/2$, so that the ordered indicators are independent fair coin flips $e_i$. A null replicate of $\Lambda$ under $\vartheta_0$ is thus exactly obtained from a vector $\mathbf e\in\{0,1\}^n$ of fair coin flips and a uniform permutation $\omega$. For $M$ replicates, the numerators are the sup-norms of the $M$ products $X^\T\mathbf u^\omega$ with the fixed matrix $X$, and the denominators are $2(n\sum_ie_i\log r_i)^{1/2}$.

\subsection*{Computing time}\label{app:2:time}
Table~\ref{tab:timing} compares the time needed to draw $M=2000$ replicates of $\Lambda$ with two schemes: the naive scheme, which at every draw simulates the null times, sorts them, reorders the rows of $X$ and rebuilds the risk-set sums, and the martingale-residual scheme above, as implemented in the R package \texttt{picreg}, with one product $X^\T\mathbf u^\omega$ per draw. The designs are fixed standardized Gaussian matrices, and each time is the median of three repetitions. Both schemes simulate the same law: their $95\%$ quantiles agree up to Monte Carlo error. The residual scheme is $29$ to $251$ times faster. The gain grows with $p$, because the naive scheme copies and sums the whole $n\times p$ matrix at every draw, whereas the residual scheme only computes $n$ residuals and one matrix--vector product with a matrix that stays in place; for large $p$ it also grows with $n$. The naive scheme is coded in R and \texttt{picreg} in C++, but an R implementation of the residual scheme runs within a factor $1.5$ of \texttt{picreg}, so that the gain is due to the algorithm rather than to the language.

\begin{table}
\centering
\small
\begin{tabular}{rrrrrcc}
\toprule
 & & \multicolumn{2}{c}{Time (seconds)} & & \multicolumn{2}{c}{$95\%$ quantile} \\
\cmidrule(lr){3-4} \cmidrule(lr){6-7}
$n$ & $p$ & Naive & Residuals & Speed-up & Naive & Residuals \\
\midrule
100 & 100 & 0.893 & 0.028 & 32 & 0.0873 & 0.0858 \\
100 & 500 & 3.71 & 0.039 & 95 & 0.0965 & 0.0967 \\
100 & 1000 & 7.94 & 0.077 & 103 & 0.0995 & 0.0996 \\
250 & 100 & 1.57 & 0.055 & 29 & 0.0505 & 0.0501 \\
250 & 500 & 6.98 & 0.089 & 78 & 0.0563 & 0.0564 \\
250 & 1000 & 16.7 & 0.085 & 197 & 0.0590 & 0.0595 \\
500 & 100 & 2.85 & 0.097 & 29 & 0.0342 & 0.0335 \\
500 & 500 & 15.8 & 0.132 & 120 & 0.0381 & 0.0379 \\
500 & 1000 & 38.5 & 0.153 & 251 & 0.0392 & 0.0390 \\
\bottomrule
\end{tabular}
\caption{Time in seconds to draw $M=2000$ null replicates of $\Lambda$ with the naive scheme and with the martingale-residual scheme of \texttt{picreg}, speed-up, and $95\%$ quantiles of the two sets of draws.}
\label{tab:timing}
\end{table}

\subsection*{Proof of Proposition~\ref{prop:mc}}
Let $F$ and $F_0$ be the distribution functions of $\Lambda$ under $\vartheta$ and under the coin-flip reference $\vartheta_0$, $\hat F_M$ the empirical distribution function of the $M$ draws, and $\rho=\sup_t|F(t)-G_n(t)|$, $\rho_0=\sup_t|F_0(t)-G_n(t)|$, which are bounded by Theorem~\ref{thm:PIC}(ii), where $G_n(t)={\rm pr}[\|Z\|_\infty\le2\{n(\log n-1)\}^{1/2}t]$ is continuous. By definition of the empirical quantile, $\hat F_M(\hat\lambda^{\rm MC}_\alpha-)\le1-\alpha\le\hat F_M(\hat\lambda^{\rm MC}_\alpha)$, and by the Dvoretzky--Kiefer--Wolfowitz inequality $\eta_M=\sup_t|\hat F_M(t)-F_0(t)|=O_{\rm pr}(M^{-1/2})$, the same bound holding for left limits. Since the jumps of $F_0$ are at most $2\rho_0$,
\begin{equation*}
    \bigl|F(\hat\lambda^{\rm MC}_\alpha)-(1-\alpha)\bigr|
    \le\sup_t|F(t)-F_0(t)|+\bigl|F_0(\hat\lambda^{\rm MC}_\alpha)-(1-\alpha)\bigr|
    \le \rho+3\rho_0+\eta_M .
\end{equation*}
Since the draws are independent of the data, ${\rm pr}_\vartheta(\Lambda>\hat\lambda^{\rm MC}_\alpha)=E\{1-F(\hat\lambda^{\rm MC}_\alpha)\}$, so that $|{\rm pr}_\vartheta(\Lambda>\hat\lambda^{\rm MC}_\alpha)-\alpha|\le\rho+3\rho_0+E(\eta_M)$ with $E(\eta_M)=O(M^{-1/2})$, and the bound follows from Theorem~\ref{thm:PIC}(ii), with $\kappa_{\vartheta_0}=0$ for the coin-flip reference.

Strictly, assumption~\ref{as1} requires end-of-study censoring, which the coin-flip mechanism does not have. The bound therefore holds for coin-flip draws censored at a late time $\tau$, with a fraction $k_\tau$ of observations still at risk, up to an additional term $12k_\tau\log(1/k_\tau)/(1-k_\tau)$ in the brackets, which is $0.56$ for $k_\tau=0.01$ and vanishes as $k_\tau\to0$.

\section{Additional simulation results}\label{app:3}
\subsection*{Implementation of the censoring mechanisms}
The five mechanisms of Section~\ref{sct:notation} are simulated as follows. Under $H_0$, in Fig.~\ref{fig:pivotality}, the event times are exponential, and the free parameter of each mechanism is calibrated once to the target censoring rate: the rate of the exponential censoring times in (i) and (v), where in (v) this rate is multiplied by $\exp(x_{i1})$; the end of study $\tau$ in (ii) and (iii); and, in (iv), the event times are shifted by one and each subject is censored with the target probability at a uniform time on $[0,1]$, so that every censored observation precedes every event. Under the alternative, in Sections~\ref{subsct:MC} and~\ref{subsct:robust}, the event times follow the exponential Cox model of Section~\ref{subsct:MC}, and the free parameter is calibrated on the simulated event times of each dataset: the rate of the exponential censoring times in (i); $\tau$ equal to the corresponding quantile of the simulated event times in (ii) and (iii); in (iv), each subject is censored with probability $w$ at a uniform time on $[0,q]$, with $q$ the $10\%$ quantile of the simulated event times, and $w$ set to reach the target rate; and in (v), the rate of the exponential censoring times is multiplied by the exponential of a noise covariate, so that event and censoring times remain independent given the covariates. Under $H_0$, the latter version of (iv) has $\kappa_\vartheta\approx-0.33$, slightly less extreme than the former, since a few early events mix with the censored observations.

\subsection*{Phase transition under the extreme censoring patterns}
\begin{figure}
\centerline{\includegraphics[width=\linewidth]{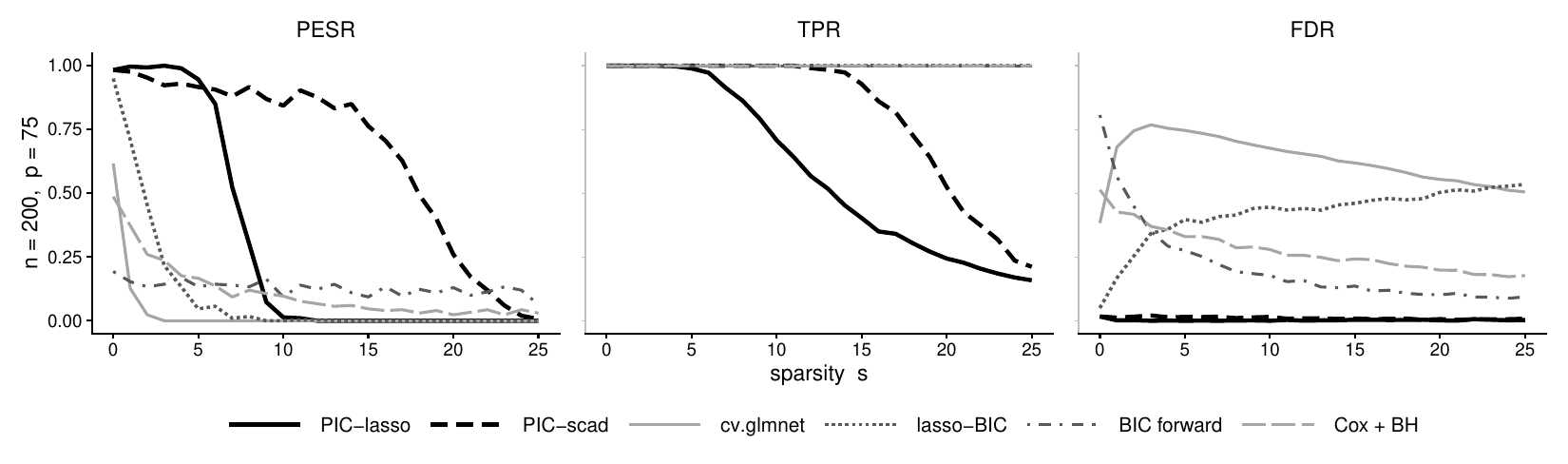}}
\centerline{\includegraphics[width=\linewidth]{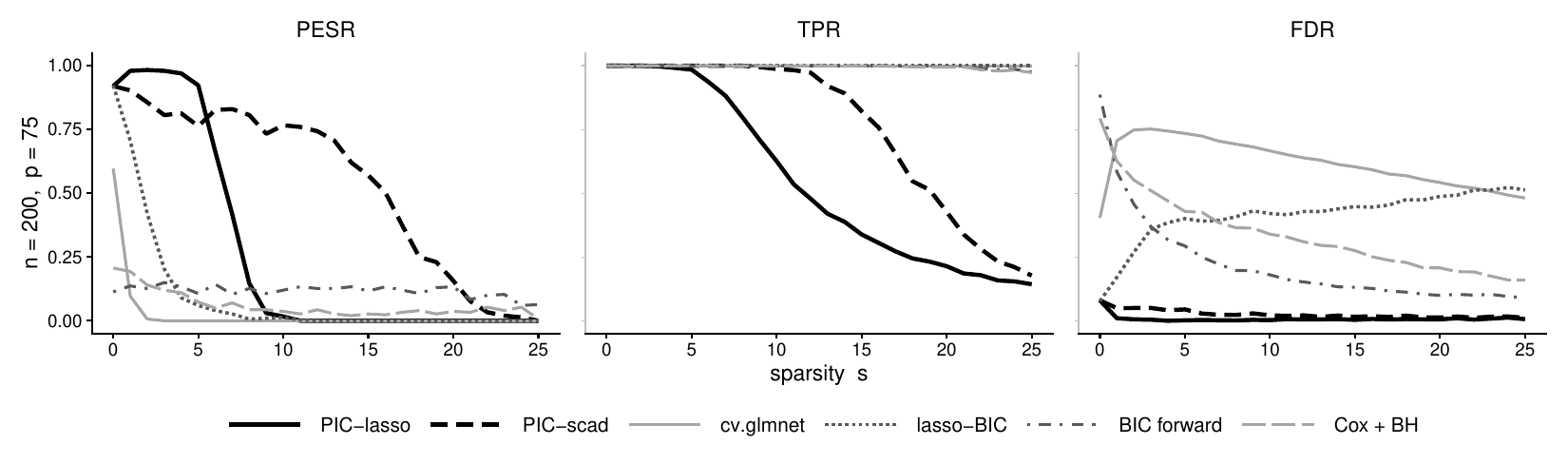}}
\caption{Phase transition of support recovery for $n=200$ and $p=75$, as in the second row of Fig.~\ref{fig:phase_transition_simu}, with about $30\%$ administrative censoring (top) and early censoring (bottom), averaged over $300$ Monte Carlo replicates.}
\label{fig:phase_transition_cens}
\end{figure}

Figure~\ref{fig:phase_transition_cens} repeats the experiment of Fig.~\ref{fig:phase_transition_simu} for $n=200$ and $p=75$, with administrative and with early censoring, about $30\%$ in both cases, and the same design, coefficients and seeds; only the censoring times differ. The case $s=0$ illustrates the direction of the error predicted by Theorem~\ref{thm:PIC}: the probability of exact support recovery of both \PIC~variants, which equals one minus the level, is $0.98$ under administrative censoring, $0.94$ under exponential censoring and $0.92$ under early censoring, that is, levels of $0.02$, $0.06$ and $0.08$, in line with Fig.~\ref{fig:pivotality}(c). For $s>0$, the curves are close to those of Fig.~\ref{fig:phase_transition_simu}. Under administrative censoring, the conservative calibration gives \texttt{PIC-scad} a slightly higher plateau of exact support recovery; under early censoring, both variants lose some power, and their phase transitions occur one unit of $s$ earlier. The competitors are hardly affected, except \texttt{Cox+BH}, whose probability of exact support recovery at $s=0$ drops from $0.49$ under administrative censoring to $0.21$ under early censoring.

\subsection*{Higher censoring rate}
Table~\ref{tab:robust60} repeats the experiment of Section~\ref{subsct:robust} with a censoring rate of about $60\%$, that is, about $80$ events for $n=200$; all other settings are unchanged.

\begin{table}
\centering
\footnotesize
\setlength{\tabcolsep}{3.4pt}
\begin{tabular}{lcccccc}
\toprule
Censoring & \texttt{PIC-lasso} & \texttt{PIC-scad} & \texttt{cv.glmnet} & \texttt{lasso-BIC} & \texttt{BIC} & \texttt{Cox+BH} \\
\midrule
\multicolumn{7}{l}{\textit{Independent design}} \\
\addlinespace[2pt]
Exponential & 4.6 / 0 / 91 & 5.1 / 2 / 100 & 23.0 / 73 / 100 & 9.1 / 39 / 100 & 8.2 / 35 / 100 & 13.7 / 51 / 85 \\
Administrative & 4.6 / 1 / 91 & 5.1 / 2 / 100 & 24.3 / 74 / 100 & 8.9 / 38 / 100 & 8.0 / 33 / 100 & 14.5 / 55 / 99 \\
Staggered & 4.5 / 0 / 89 & 5.0 / 1 / 100 & 20.4 / 73 / 100 & 9.7 / 42 / 100 & 8.2 / 35 / 100 & 14.2 / 55 / 100 \\
Early & 4.1 / 1 / 81 & 5.3 / 5 / 100 & 22.6 / 71 / 100 & 11.1 / 43 / 100 & 8.7 / 38 / 100 & 20.6 / 53 / 65 \\
Covariate & 4.6 / 0 / 92 & 5.1 / 2 / 100 & 23.2 / 73 / 100 & 8.6 / 38 / 100 & 8.1 / 33 / 100 & 13.2 / 50 / 87 \\
\midrule
\multicolumn{7}{l}{\textit{Toeplitz design, $\rho=0.5$}} \\
\addlinespace[2pt]
Exponential & 4.7 / 3 / 90 & 5.1 / 3 / 99 & 23.6 / 72 / 100 & 10.5 / 47 / 100 & 7.8 / 32 / 100 & 13.1 / 49 / 82 \\
Administrative & 4.5 / 3 / 87 & 5.1 / 1 / 100 & 22.8 / 73 / 100 & 10.5 / 47 / 100 & 7.6 / 30 / 100 & 16.0 / 59 / 98 \\
Staggered & 4.6 / 3 / 89 & 5.1 / 2 / 100 & 23.4 / 74 / 100 & 10.8 / 48 / 100 & 7.9 / 33 / 100 & 14.7 / 58 / 98 \\
Early & 4.3 / 5 / 80 & 5.3 / 6 / 98 & 19.9 / 69 / 100 & 13.7 / 48 / 99 & 7.7 / 32 / 100 & 19.8 / 53 / 62 \\
Covariate & 4.8 / 3 / 91 & 5.2 / 3 / 100 & 26.5 / 75 / 100 & 10.1 / 46 / 100 & 7.7 / 31 / 100 & 12.9 / 50 / 83 \\
\midrule
\multicolumn{7}{l}{\textit{Toeplitz design, $\rho=0.8$}} \\
\addlinespace[2pt]
Exponential & 5.2 / 24 / 77 & 5.0 / 11 / 88 & 21.8 / 74 / 99 & 12.7 / 59 / 97 & 7.4 / 31 / 98 & 13.0 / 55 / 70 \\
Administrative & 5.1 / 19 / 79 & 4.9 / 8 / 89 & 18.0 / 70 / 98 & 11.8 / 55 / 96 & 7.0 / 27 / 98 & 12.3 / 57 / 78 \\
Staggered & 5.2 / 23 / 78 & 4.9 / 8 / 90 & 20.9 / 72 / 99 & 12.9 / 58 / 98 & 7.4 / 29 / 99 & 13.9 / 59 / 77 \\
Early & 5.1 / 24 / 74 & 5.2 / 17 / 86 & 16.3 / 69 / 99 & 13.5 / 58 / 98 & 7.7 / 32 / 98 & 22.5 / 61 / 55 \\
Covariate & 5.4 / 24 / 79 & 4.9 / 12 / 86 & 22.6 / 72 / 99 & 13.0 / 59 / 98 & 7.5 / 31 / 97 & 12.6 / 55 / 66 \\
\bottomrule
\end{tabular}
\caption{As Table~\ref{tab:robust}, with about $60\%$ censoring: $\hat s$ / FDR / TPR at $s=5$, for $n=200$ and $p=75$, averaged over $100$ runs.}
\label{tab:robust60}
\end{table}

\bibliographystyle{plainnat}
\bibliography{article_v2}
\end{document}